\documentclass[10pt,final,twocolumn,letterpaper,pagenumbers]{article}
\usepackage{style}               
\usepackage{url}
\usepackage{graphicx}
\usepackage{tabularx}
\usepackage{comment}
\usepackage{makecell}
\usepackage{pifont}
\usepackage{multirow}
\usepackage[misc]{ifsym}
\usepackage{afterpage}
\usepackage{setspace}
\usepackage{breakcites}
\usepackage{seqsplit}
\usepackage{multirow}
\usepackage{booktabs}
\usepackage{times}
\usepackage{epsfig}
\usepackage{graphicx}
\usepackage{amsmath}
\usepackage{amssymb}
\usepackage{float}
\usepackage[symbol]{footmisc}
\usepackage[title]{appendix}
\usepackage{lipsum}

\usepackage{hyphenat}
\usepackage{soul,color}
\usepackage{amsopn}

\definecolor{Red}{cmyk}{0,1,1,0}
\definecolor{Green}{cmyk}{1,0,1,0}
\definecolor{Cyan}{cmyk}{1,0,0,0}
\definecolor{Purple}{cmyk}{0.45,0.86,0,0}
\definecolor{Rosolic}{cmyk}{0.00,1.00,0.50,0}
\definecolor{Blue}{cmyk}{1.00,1.00,0.00,0}
\definecolor{BlueViolet}{cmyk}{0.86,0.91,0,0.04}
\definecolor{NavyBlue}{cmyk}{0.94,0.54,0,0}

\newcommand{\hidden}[1]{{\color{NavyBlue}}}

\newcommand{\myparagraph}[1]{\vspace{0.4em}\noindent\textbf{#1}}

\newcommand{\orange}[1]{\textcolor[RGB]{255,128,0}{#1}}

\definecolor{cvprblue}{rgb}{0.21,0.49,0.74}
\usepackage[pagebackref,breaklinks,colorlinks,citecolor=cvprblue]{hyperref}

\usepackage[capitalize]{cleveref}
\crefname{section}{Sec.}{Secs.}
\Crefname{section}{Section}{Sections}
\Crefname{table}{Table}{Tables}
\crefname{table}{Tab.}{Tabs.}

\begin{document}


\title{M3DBench: Let's Instruct Large Models with Multi-modal 3D Prompts}

\author{
    Mingsheng Li$^{1}$ \quad
    Xin Chen$^{2,*}$ \quad
    Chi Zhang$^{2}$ \quad
    Sijin Chen$^{1}$ \quad
    Hongyuan Zhu$^{3}$ \quad
    \\
    Fukun Yin$^{1}$ \quad
    Gang Yu$^{2}$ \quad
    Tao Chen$^{1,\dagger}$
    \\
    {\normalsize $^{1}$Fudan University} \quad
    {\normalsize $^{2}$Tencent PCG} \quad \\
    {\normalsize $^{3}$Institute for Infocomm Research (I$^2$R) \&
     Centre for Frontier AI Research (CFAR), A*STAR, Singapore}
     \\
     \tt \small \textbf{\href{https://github.com/OpenM3D/M3DBench/}{https://github.com/OpenM3D/M3DBench}}
    \\
    {\small $^*$ Project Lead \ \ \ \ \ \ \ \small $^{\dagger}$ Corresponding Author}
}




\twocolumn[{
    \renewcommand\twocolumn[1][]{#1}
    \maketitle
    \begin{center}
        \captionsetup{type=figure}
        \includegraphics[width=\textwidth]{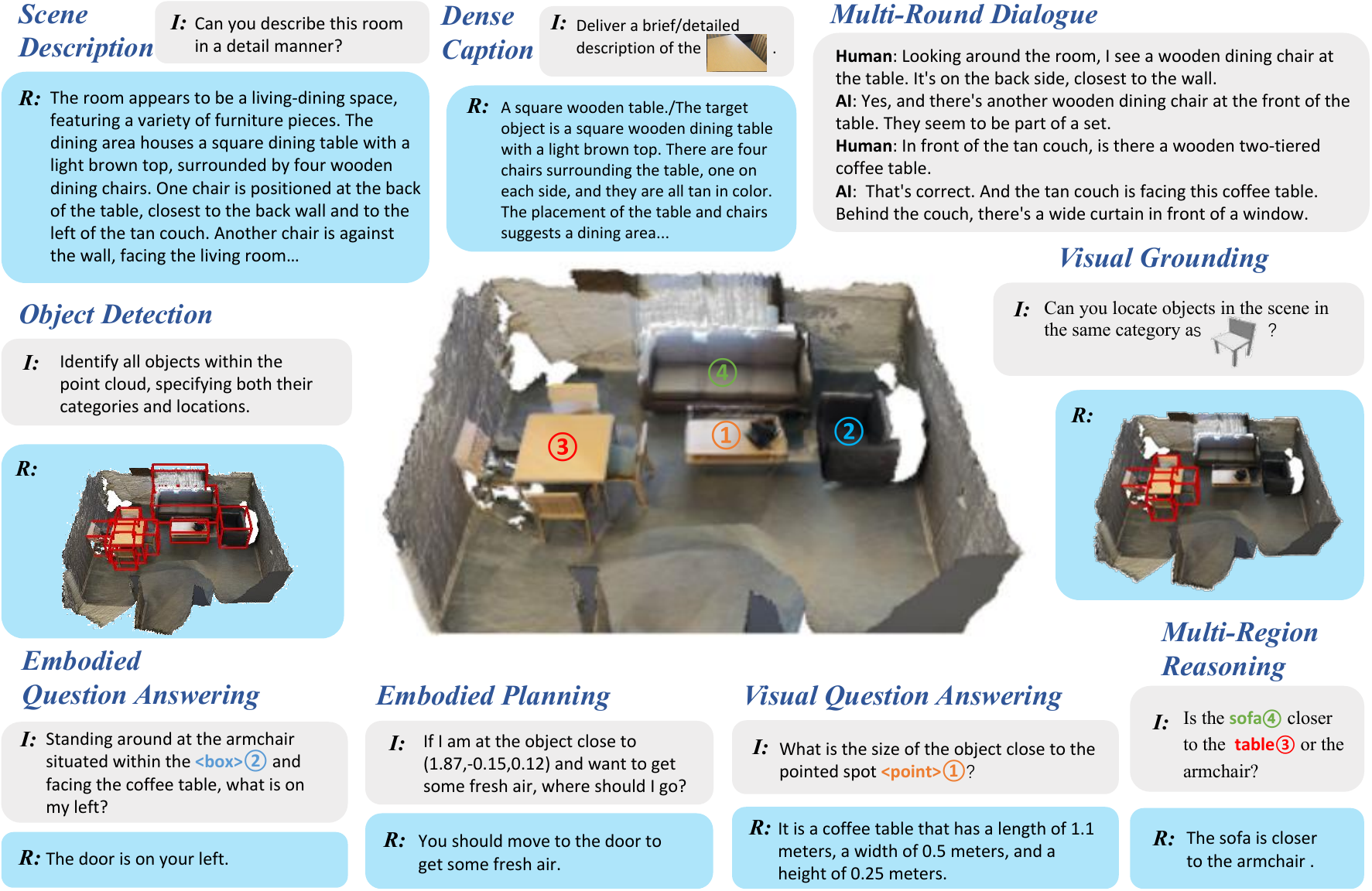}
        \caption{
            \textbf{Examples from M3DBench, which encompasses a variety of 3D-centric tasks.} The dataset supports multi-modal
            instructions that interleave text with visual prompts and covers a variety of fundamental abilities in real-world 3D environments, such as visual perception, scene understanding, spatial reasoning, navigation, and planning.
        }
        \label{fig:teaser}
        \vspace{4pt}
    \end{center}
}]

\begin{abstract}

Recently, 3D understanding has become popular to facilitate autonomous agents to perform further decision-making. However, existing 3D datasets and methods are often limited to specific tasks. On the other hand, recent progress in Large Language Models (LLMs) and Multi-modal Language Models (MLMs) have demonstrated exceptional general language and imagery tasking performance. Therefore, it is interesting to unlock MLM's potential to be 3D generalist for wider tasks.  
However, current MLMs' research has been less focused on 3D tasks due to a lack of large-scale 3D instruction-following datasets. 
In this work, we introduce a comprehensive 3D instruction-following dataset called M3DBench, which possesses the following characteristics: 1) It supports \textbf{general multi-modal instructions} interleaved with text, images, 3D objects, and other visual prompts. 2) It unifies \textbf{diverse 3D tasks at both region and scene levels}, covering a variety of \textbf{fundamental abilities} in real-world 3D environments. 3) It is a large-scale 3D instruction-following dataset with \textbf{over 320k instruction-response pairs}. 
Furthermore, we establish a new benchmark for assessing the performance of large models in understanding multi-modal 3D prompts.
Extensive experiments demonstrate the effectiveness of our dataset and  baseline, supporting general 3D-centric tasks, which can inspire future research. 

\end{abstract}


\section{Introduction}
\label{sec:intro}

\hspace{1em}
The past year has witnessed remarkable success of \textbf{L}arge \textbf{L}anguage \textbf{M}odels(LLMs) families\cite{t5/raffel2020exploring, llama/touvron2023llama, vicuna/chiang2023vicuna, chatgpt/schulman2022chatgpt} in addressing various natural language processing tasks through general instruction tuning~\cite{instruction-tuning/ouyang2022training}.
%
%
\textbf{M}ulti-modal \textbf{L}anguage \textbf{M}odels (MLMs), such as Flamingo~\cite{flamingo/alayrac2022flamingo}, BLIP-2~\cite{blip-2/li2023blip}, LLaVA~\cite{llava/liu2023visual} have progressed various visual comprehension and reasoning tasks on 2D domain, including visual captioning~\cite{llm_caption1/bianco2023improving, llm_caption2/rotstein2023fusecap, llm_caption3/wang2023caption}, dialogue~\cite{llm_dialog/chen2023x} and question-answering~\cite{llm_vqa1/hu2023bliva, llm_vqa2/guo2023images}. 
To unlock the full potential of these MLMs, it is essential to curate a well-constructed instruction-following dataset~\cite{llava/liu2023visual,mimic-it/li2023mimic} that covers diverse vision language (VL) tasks, which empowers the models to handle these tasks without extensive modifications to the architecture. 
However, current research on MLMs has predominantly overlooked 3D visual and a comprehensive dataset for 3D instruction tunning is missing due to the daunting workload of collecting instructions in ambiguous and cluttered 3D environments. 

%
%
Previous works have made efforts to construct datasets for specialized 3D task, such as object detection~\cite{scannet/dai2017scannet,sunrgbd/song2015sun}, visual grounding~\cite{referit3d/achlioptas2020referit3d,scanrefer/chen2020scanrefer}, dense captioning~\cite{referit3d/achlioptas2020referit3d,scanrefer/chen2020scanrefer}, VQA~\cite{scanqa/azuma2022scanqa,clevr3d/yan2021clevr3d}, and navigation~\cite{r2r/anderson2018vision}.
%
Consequently, most of the models~\cite{votenet/qi2019deep,3detr/misra2021end,3djcg/cai20223djcg,dnet/chen2022d,unit3d/chen2023unit3d,scanqa/azuma2022scanqa} are specialist in only one or two of these tasks, potentially limiting their adaptability across various applications.
Works such as LAMM~\cite{lamm/yin2023lamm}, 3D-LLM~\cite{3d-llm/hong20233d}, and Chat-3D~\cite{chat-3d/wang2023chat} have made preliminary attempts in constructing 3D instruction-following datasets, achieving inspiring results.
However, the range of visual tasks covered by these datasets is relatively \textit{limited}, which constrains their effectiveness under diverse scenarios.
These datasets primarily focus on language-only instructions, posing challenges in identifying specific object within a scene.  For example, there might be multiple instances of ``wooden chair'' in a scene, yet the language prompt pertaining to a specific wooden chair might result in \textit{ambiguity}.
Furthermore, the lack of a comprehensive evaluation \textit{benchmark} poses challenges in accurately assessing the capability of large models on 3D-centric tasks. 
Current works, such as LAMM\cite{lamm/yin2023lamm}, primarily evaluate model's performance on previous benchmarks that are not designed for assessing MLMs with open-form output~\cite{3d-llm/hong20233d}.

\begin{table*}[t]
    \centering
    \resizebox{\linewidth}{!}{
    \begin{tabular}{ccccccccccccccccccccccc}
    \toprule
        \multirow{3}{*}{Dataset} & \multicolumn{2}{c}{Statistics}  &                                                      & \multicolumn{6}{c}{Instruction}    &    & \multicolumn{2}{c}{Perception} &  & \multicolumn{6}{c}{Understanding and Reasoning}                                                             &    & \multicolumn{2}{c}{Planning}              \\ \cline{2-3} \cline{5-10} \cline{12-13} \cline{15-20} \cline{22-23} 
    
                               &  {\begin{tabular}[c]{@{}c@{}}\#Instruction- \\response pairs\end{tabular}}   & {\begin{tabular}[c]{@{}c@{}}\#Average length of \\ instrucion / response\end{tabular}} & &Text & Coord & Point & Box  & Image &  {\begin{tabular}[c]{@{}c@{}}3D \\Object\end{tabular}}& &{\begin{tabular}[c]{@{}c@{}}Object \\ Detection\end{tabular}}    & {\begin{tabular}[c]{@{}c@{}}Visual \\ Grounding\end{tabular}}  & & {\begin{tabular}[c]{@{}c@{}}Dense \\ Caption\end{tabular}} & {\begin{tabular}[c]{@{}c@{}}Visual \\ Question\\Answering\end{tabular}} & {\begin{tabular}[c]{@{}c@{}}Embodied \\ Question\\Answering\end{tabular}} & {\begin{tabular}[c]{@{}c@{}}Multi- \\ region \\Reasoning\end{tabular}} & {\begin{tabular}[c]{@{}c@{}}Scene \\ Description\end{tabular}}  & {\begin{tabular}[c]{@{}c@{}}Multi- \\ round\\Dialogue\end{tabular}} & & {\begin{tabular}[c]{@{}c@{}}Embodied\\Planning\end{tabular}} & {\begin{tabular}[c]{@{}c@{}}Vision- \\ Language\\Navigation\end{tabular}}\\ \hline

    Nr3D~\cite{referit3d/achlioptas2020referit3d}       & - &  -  &   & $\checkmark$   & $\times$     & $\times$     & $\times$     & $\times$   & $\times$     &    & $\times$            & $\checkmark$      &            & $\checkmark$              & $\times$           & $\times$               & $\times$                 & $\times$   & $\times$                 &  & $\times$                          & $\times$                 \\
    ScanRefer~\cite{scanrefer/chen2020scanrefer}     & - &  -  &  & $\checkmark$    & $\times$     & $\times$     & $\times$     & $\times$   & $\times$    &     & $\times$            & $\checkmark$        &          & $\checkmark$              & $\times$           & $\times$               & $\times$                 & $\times$   & $\times$             &       & $\times$                          & $\times$                 \\
    ScanQA~\cite{scanqa/azuma2022scanqa}       &25K & 8.77 / 2.42     &  & $\checkmark$    & $\times$     & $\times$     & $\times$     & $\times$   & $\times$        &  & $\times$            & $\times$   &                & $\times$      & $\checkmark$          & $\times$           & $\times$               & $\times$                  & $\times$          &           & $\times$                          & $\times$                 \\
    SQA3D~\cite{sqa3d/ma2022sqa3d}        & 26K & 10.49 / 1.10      &   & $\checkmark$    & $\times$     & $\times$     & $\times$     & $\times$   & $\times$     &     & $\times$            & $\times$ &                  & $\times$      & $\checkmark$          & $\checkmark$           & $\times$               & $\times$                  & $\times$          &           & $\checkmark$                          & $\times$                 \\ \hline

    ScanScribe~\cite{scanscribe/zhu20233d}    &  - & -    &        & $\checkmark$    & $\times$    & $\times$     & $\times$     & $\times$   & $\times$     &    & -            & -     &             & -              & -           & -               & -                 & -   & -                    && -                          & -                 \\
    LAMM-3D~\cite{lamm/yin2023lamm}     & 10K  & 13.88 / 119.34    &     & $\checkmark$    & $\times$     & $\times$     & $\times$     & $\times$   & $\times$   &      & $\checkmark$            & $\times$          &        & $\times$     & $\checkmark$             & $\times$           & $\times$               & $\checkmark$                & $\checkmark$              &      & $\times$                          & $\times$                 \\
    3DLLM~\cite{3d-llm/hong20233d}      & 202K & 43.80 / 8.11    &       & $\checkmark$    & $\checkmark$     & $\times$     & $\times$     & $\times$   & $\times$     &    & $\times$            & $\checkmark$         &         & $\checkmark$   & $\checkmark$             & $\checkmark$           & $\times$               & $\checkmark$                  & $\checkmark$                 &   & $\checkmark$                          & $\checkmark$                 \\
    Chat-3D~\cite{chat-3d/wang2023chat}        & 57K &   9.11 / 48.75     &          & $\checkmark$    & $\times$     & $\times$     & $\times$     & $\times$   & $\times$     &    & $\times$            & $\times$     &          & $\checkmark$    & $\times$          & $\times$           & $\times$               & $\times$                  &   $\checkmark$      &            & $\times$   & $\times$                                            \\ \hline
    M3DBench     &  327K &   24.79 / 18.48       &                 & $\checkmark$    & $\checkmark$     & $\checkmark$     & $\checkmark$     & $\checkmark$   & $\checkmark$   &      & $\checkmark$            & $\checkmark$     &             & $\checkmark$              & $\checkmark$           & $\checkmark$               & $\checkmark$                 & $\checkmark$   & $\checkmark$            &        & $\checkmark$                          & $\checkmark$                \\
    \bottomrule
    \end{tabular} 
    }
    \caption{
        \textbf{Comparison between M3DBench and other 3D VL datasets as well as 3D instruction datasets.} M3DBench has the following characteristics: \textbf{1)} A \textbf{comprehensive} instruction-following dataset tailored for 3D scenes. \textbf{2)} Supporting \textbf{multi-modal instructions} that interleave text, coordinate, image, 3D object, and so on. \textbf{3)} Encompassing \textbf{diverse 3D visual-centric tasks} that span a variety of \textbf{fundamental abilities} in real-world 3D environments, such as visual perception, scene understanding, spatial reasoning, navigation, and planning.
    }
    \label{tab:comparasion_dataset}
\end{table*}



In this paper, we introduce a comprehensive 3D instruction-following dataset called M3DBench, serving as the foundation for developing a versatile and practical general-purpose assistant in the real-world 3D environment.
Our dataset comprises a variety of 3D vision-centric tasks at both object and scene levels and over 320K 3D instruction-response pairs, covering fundamental capabilities such as visual perception, scene understanding, spatial reasoning, and embodied planning, VL navigation, as depicted in \cref{tab:comparasion_dataset}. 
%
Furthermore, to tackle the challenge of ambiguity in language-only instructions, we interleave text instructions with other prompts that provide rich clues about instances in the scene, such as numerical coordinates, pointed region, image, 3D object (as shown in ~\cref{fig:teaser}) in M3DBench, to enhance the capabilities in comprehending different granularity, diversity and interactivity concepts (such as ``the pointed region'' or ``find the \textit{$\langle \text{image of a whiteboard} \rangle$} in the room'') in the multi-modal instructions. 
%




    

To evaluate the effectiveness of M3DBench, we develop a simple yet effective baseline model capable of processing interleaved multi-modal instructions, consisting of three components: scene perceiver, multi-modal instruction encoder, and LLM decoder.
Furthermore, we develop a comprehensive benchmark aimed at systematically assessing various capabilities of 3D MLMs across multiple dimensions with multi-modal instructions .
The evaluation benchmark comprises approximately 1.5K instruction-response pairs, encompassing both region-level and scene-level tasks, such as object localization, scene description, multi-round dialogues, embodied planning, among others.
Each instance comprises an instruction, a corresponding 3D scene, and a human-validated response. 
%
We will release M3DBench dataset, code, and evaluation strategies to accelerate future research on 3D MLMs.




%
%
%




To summarize, our contributions are listed as following:
\begin{itemize} 
\setlength\itemsep{0em}
    \item We introduce a large-scale 3D instruction-following dataset that unifies diverse region-level and scene-level 3D-centric tasks, focusing on scene perception, understanding, reasoning, and planning. 

    \item We present a interleaved multi-modal instruction formula designed to enhance the granularity, diversity and interactivity of generated instructions.

    \item We establish a comprehensive benchmark for evaluating the capabilities of MLMs within 3D scenarios. Extensive experiments demonstrate the effectiveness of both the dataset and the baseline.

\end{itemize}

\section{Related Work}
\label{sec:related}

\myparagraph{Multi-modal Datasets and 3D Benchmarks.} 
%
The progress of MLMs~\cite{clip/radford2021learning, vilt/kim2021vilt, blip/li2022blip, blip-2/li2023blip} has been greatly accelerated by the availability of large-scale image-text data, such as MS COCO Caption~\cite{coco-caption/chen2015microsoft}, Visual Genome~\cite{vg/krishna2017visual}, LAION-5B~\cite{laion/schuhmann2022laion}.
%
%
In order to improve models' comprehension of human instructions in visual tasks, several visual instruction following datasets~\cite{llava/liu2023visual,mimic-it/li2023mimic,lamm/yin2023lamm, instructblip/dai2023instructblip} have been proposed.
Additionally, while numerous studies in the field of 3D have presented benchmark datasets for visual grounding~\cite{referit3d/achlioptas2020referit3d, scanrefer/chen2020scanrefer}, dense captioning~\cite{referit3d/achlioptas2020referit3d, scanrefer/chen2020scanrefer}, and visual question answering~\cite{scanqa/azuma2022scanqa,sqa3d/ma2022sqa3d}, these datasets are limited to specific tasks.
In this paper, we propose a comprehensive dataset that supports interleaved multi-modal instructions and covers various 3D-centric tasks, including visual grounding, dense caption, embodied question answering, multi-region reasoning, scene description, multi-round dialogue, and so on.
%
%
Refer to \cref{tab:comparasion_dataset} for a detailed comparison between our dataset and other 3D VL datasets~\cite{referit3d/achlioptas2020referit3d,scanrefer/chen2020scanrefer,scanqa/azuma2022scanqa,sqa3d/ma2022sqa3d} as well as exiting 3D visual instruction datasets~\cite{lamm/yin2023lamm,3d-llm/hong20233d,chat-3d/wang2023chat}.
Furthermore, rather than providing demonstrations only, we evaluate diverse tasks with quantitative results.

\myparagraph{Multi-modal Foundation Models.}
%
%
With the triumph of LLMs~\cite{gpt-3/brown2020language, opt/zhang2022opt, t5/raffel2020exploring, llama/touvron2023llama, vicuna/chiang2023vicuna}, recent studies~\cite{flamingo/alayrac2022flamingo, blip-2/li2023blip, llava/liu2023visual, otter/li2023otter} start to explore Vision Language Models (VLMs), extending the capabilities of LLMs in solving diverse visual-related tasks.
Early attempts include Flamingo~\cite{flamingo/alayrac2022flamingo}, which incorporates visual features through gated cross-attention dense blocks, and BLIP-2~\cite{blip-2/li2023blip}, which uses a Q-former as a bridge to reconcile the modality gap between the frozen image encoder and LLMs.
In order to enhance the VLMs' comprehension of human instructions, several visual instruction tuning methods~\cite{llava/liu2023visual, otter/li2023otter} have been proposed.
Addressing the adaptation of LLMs to 3D-related tasks, LAMM~\cite{lamm/yin2023lamm} uses a simple projection layer to connect the 3d encoder and LLM. 
3D-LLM~\cite{3d-llm/hong20233d} utilizes point clouds and text instructions as input, leveraging 2D VLMs as backbones. 
%
However, prior works that attempt to integrate the 3D world into MFMs have exhibited limitations in handling interleaved multi-modal instructions and accomplishing various tasks.
In this work, we propose to improve the abilities of MFMs in addressing diverse 3D-centric tasks and handling interleaved multi-modal instructions with on a comprehensive 3D instruction-following dataset.

\myparagraph{3D Vision-language Learning.}
%
%
Recently, there has been growing interest in 3D VL learning. 
While various 3D representations exist, including voxels, point clouds, and neural fields, previous works have primarily focused on point cloud-text data. 
Among those, 3D dense captioning~\cite{dc_scan2cap/chen2021scan2cap, x-trans/yuan2022x} aims to generate description of target object within a 3D scene, while 3D visual grounding~\cite{sat/yang2021sat,3dvg/zhao20213dvg,instancerefer/yuan2021instancerefer} involves identifying object in a scene based on textual description.
In 3D question answering~\cite{scanqa/azuma2022scanqa,clevr3d/yan2021clevr3d}, models are required to answer questions based on the visual information. 
Although these works have achieved impressive results in connecting 3D vision and language, they heavily rely on task-specific model design.
%
In contrast, we develop a unified baseline model capable of decoding multiple 3D-related tasks without the need for specific model designs. Furthermore, we establish a comprehensive benchmark to assess the model's performance across various tasks.

\section{Multi-modal Instruction Dataset}
\label{sec:dataset}

\hspace{1em}
We introduce the strategy for constructing the multi-modal 3D instruction dataset (details in ~\cref{sec:dataset construct}), along with the design formula for interleaved multi-modal instructions (details in ~\cref{sec:interleaved}).
We then detail the tasks at both the region-level and scene-level covered by the dataset in ~\cref{sec:dataset task}, followed by a statistical and analytical examination of the dataset in ~\cref{dataset analysis}.
%
%


\subsection{Dataset Construction}
\label{sec:dataset construct}
\hspace{1em}
To construct a comprehensive 3D multi-modal instruction-following dataset, we utilize existing datasets~\cite{scannet/dai2017scannet,scanrefer/chen2020scanrefer,referit3d/achlioptas2020referit3d,phraserefer/yuan2022toward,mp3d/chang2017matterport3d,nav/krantz2020beyond, imagenet/krizhevsky2012imagenet,shapenet/chang2015shapenet} for 3D-only~\cite{votenet/qi2019deep,vg_instancerefer/yuan2021instancerefer} and 3D-language tasks~\cite{dc_scan2cap/chen2021scan2cap,scanqa/azuma2022scanqa}, and collect extensive instruction-response data by prompting LLMs. 
For 3D-only tasks, like object detection and visual grounding, we designed instruction and response templates corresponding to specific tasks. 
Specifically, we collect instructions for 3D-only tasks by providing task descriptions and specifying the desired output format. Responses interleave object coordinates (follow the specified output format in the instructions) and text.
For the 3D-language tasks, such as dialogue and question answering, we provided object attributes, textual descriptions, and manually written task construction instructions as well as few-shot in-context learning examples to the GPT-API~\cite{chatgpt/schulman2022chatgpt,gpt-4/openai2023gpt} to generate task-specific instruction data. 

Although most responses generated by GPT-API~\cite{chatgpt/schulman2022chatgpt,gpt-4/openai2023gpt} are of high quality, some were irrelevant to the instruction. 
For instance, certain responses may refer to information derived from the provided textual descriptions.
To improve the quality of the 3D instruction-following data, we employ pattern matching with specific keywords to filter out such responses. 
%

\subsection{Interleaved Multi-modal Instruction}
\label{sec:interleaved}
\hspace{1em}We design four types of visual prompts, namely, point-level prompt (user click), box-level prompt (pointed region), image prompt, and 3D object prompt.
For point-level prompt, we sample points near the specified region of the scene and randomly select one. 
The box-level prompts are derived from the ground truth bounding boxes in the 3D scene.
Image prompts consist of corresponding image regions with 3D scenes, publicly available image data\cite{imagenet/krizhevsky2012imagenet}, and synthetic images\cite{sdxl/podell2023sdxl}.
Regarding 3D objects, we select instances from 3D scenes with ground truth per-point annotations, additionally collecting objects from \cite{shapenet/chang2015shapenet}.
Furthermore, we provide specific descriptions, such as ``in the pointed region'' in instructions to guide the large model to identify visual prompts within the scene.
Finally, an interleaved multi-modal instruction $I$ can be defined as an ordered sequence composed of text and visual prompts, represented as $I = [x^1, x^2, \dots, x^{M}]$, where each element $x^i$ in $\{text, point, box, image, object_{3d} \}$.
Additional details can be found in the supplementary materials.




\subsection{Task Coverage} 
\label{sec:dataset task}

\hspace{1em}
Our dataset introduces a unified \textit{instruction-response} format to cover diverse 3D-centric tasks, encompassing essential capabilities ranging from visual perception and understanding to reasoning and planning (detailed in ~\cref{tab:comparasion_dataset}).

\subsubsection{Visual Perception} 

\myparagraph{Object Detection(OD)} aims at identifying and locating all the objects of interest in a point cloud~\cite{od_pointformer/pan20213d,od_3detr/misra2021end}.
Here, we transform the classic OD task into an instruction-following format by providing task descriptions and specifying the desired output format.
Following LAMM~\cite{lamm/yin2023lamm}, we manually design a set of instruction-response templates with placeholders, and each instruction includes the expected output format. 
The instruction and response templates can be found in the supplementary.

\myparagraph{Visual Grounding(VG)} involves identifying the target object in the scene based on a natural language referring expression~\cite{vg_instancerefer/yuan2021instancerefer, vg_eda/wu2023eda}. 
In M3DBench, we expand the task format of VG. 
Specifically, our description information for querying extends beyond textual input and includes various visual prompts, such as coordinate, clicked point, image, 3D object, and so on. 
Moreover, our output is not limited to locating a single target object but can also involve finding objects belonging to the same category.

\subsubsection{Scene Understanding and Reasoning}

\myparagraph{Dense Caption(DC)} requires a model to generate natural language descriptions for each object~\cite{dc_scan2cap/chen2021scan2cap, dc_vote2cap/chen2023end}. 
However, existing DC datasets like ScanRefer~\cite{scanrefer/chen2020scanrefer} and Nr3D~\cite{referit3d/achlioptas2020referit3d} provide only short captions. 
In M3DBench, we reconstruct the DC datasets and introduce terms like \textit{brief} or \textit{detailed} in instruction to generate either concise title or detailed description for the object, which allows for better control over the granularity of the generated caption.
The instruction templates can be found in the supplementary.

\myparagraph{Visual Question Answering(VQA)} is a task that requires the model to correctly answer a given question based on the information present in a visual scene~\cite{scanqa/azuma2022scanqa,qa_method1/parelli2023clip}. 
In this work, we curate a collection of free-form, open-ended question-answer pairs using publicly available 3D-language datasets.
These VQA pairs cover various aspects at both the object level and scene level, including instance locations and attributes, object counts, room functions, and more. 
%

\begin{figure*}[htbp]
	\centering
	\includegraphics[width=\linewidth]{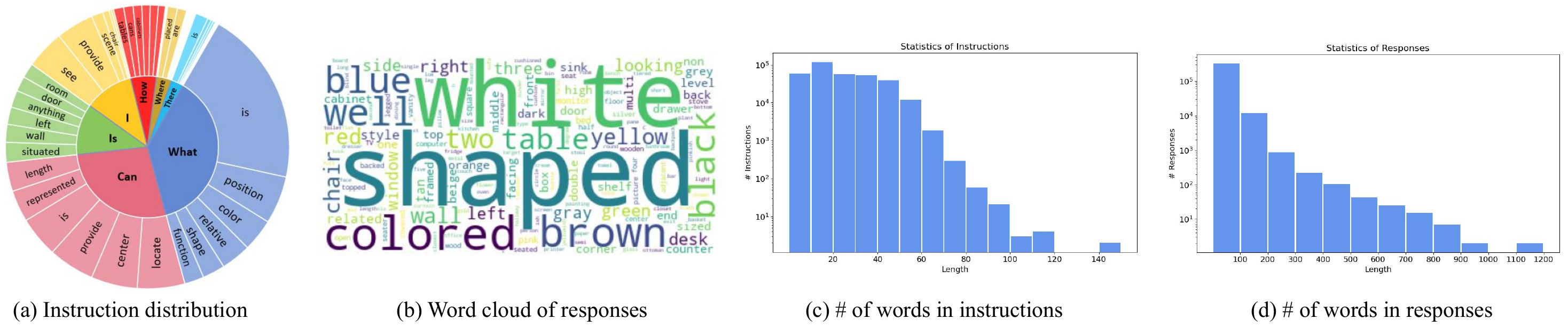}
    \setlength{\abovecaptionskip}{-0.4cm}
        \vspace{7pt}
	\caption{
        \textbf{The statistics of the M3DBench.} (a) The distribution of instructions based on the first word, where the inner circle of the graph represents the frequency of the first word's occurrence, and the outer circle shows the frequency of verbs and nouns appearing in the instructions corresponding to that first word. (b) The word cloud of responses. (c) The distribution of instruction length. (d) The distribution of response length. 
    }
	\label{fig:statistics}
\end{figure*}

\myparagraph{Embodied Question Answering(EQA).} 
%
Unlike traditional VQA tasks~\cite{scanqa/azuma2022scanqa,qa_method1/parelli2023clip} that primarily focus on answering questions related to global information, EQA requires the agent to first comprehend and analyze the surrounding environment to answer questions under that situation~\cite{sqa3d/ma2022sqa3d}.
%
%
%
To collect instruction-following data for EQA, we start by randomly selecting a location within the scene and choosing to face a nearby object for reference direction, and then prompt GPT-4 to generate EQA pairs based on the given situation and text information.

\myparagraph{Multi-region Reasoning(MR).} 
Datasets such as DC~\cite{scanrefer/chen2020scanrefer,referit3d/achlioptas2020referit3d} facilitate understanding and reasoning for individual objects.
However, reasoning between distinct regions is often overlooked. 
For instance, inquiries about the spatial relationship between $\langle \text{region 1} \rangle$ and $\langle \text{region 2} \rangle$.
Here, we introduce MR, which is designed to enhance fine-grained comprehension of multiple regions of interest. 
Our methodology involves feeding object location, descriptions~\cite{phraserefer/yuan2022toward}, few-shot learning examples, and language instructions to GPT-4 to obtain corresponding responses.

\myparagraph{Scene Description(SD).} 
Unlike DC~\cite{dc_scan2cap/chen2021scan2cap, dc_vote2cap/chen2023end}, which generates a caption for each object, SD focuses on producing descriptions of the entire scene, extending the descriptive ability of MLMs from the region level to the scene level.
To construct the instruction-following data for SD, we extract 3D bounding box annotations from ScanNet~\cite{scannet/dai2017scannet} and dense captions from the 3D VL datasets~\cite{scanrefer/chen2020scanrefer,referit3d/achlioptas2020referit3d} as data sources.
By prompting the GPT-4, we can generate detailed descriptions for each scene.

\myparagraph{Multi-round Dialogue(MD).} 
To construct MDs, we make use of 3D VL datasets and follow a similar approach to that used in LLAVA~\cite{llava/liu2023visual}. 
During this process, we prompt GPT-4 to generate MDs in a self-questioning and self-answering format, taking advantage of coordinate information and language descriptions from~\cite{referit3d/achlioptas2020referit3d,scanrefer/chen2020scanrefer}.

\subsubsection{Planning and Navigation}
\myparagraph{Embodied Planning(EP).} 
Unlike EQA, which primarily focuses on answering questions, EP requires agents to possess planning and decision-making capabilities. 
Specifically, the agent needs to perceive the environment, understand user's intentions, and generate appropriate action instructions to achieve predefined goals~\cite{3d-llm/hong20233d}.
%
%

\myparagraph{Vision Language Navigation(NLV)} require an agent to navigate and move in a real-world 3D environment based on human language instructions. 
We leverage annotations from existing 3D-language navigation tasks~\cite{nav/krantz2020beyond} and transform them into an instruction-following format.
%
Instructions are expressed in natural language, while the corresponding response is a trajectory formed by points in space.

\begin{figure*}[t]
	\centering
	\includegraphics[width=0.9\linewidth]{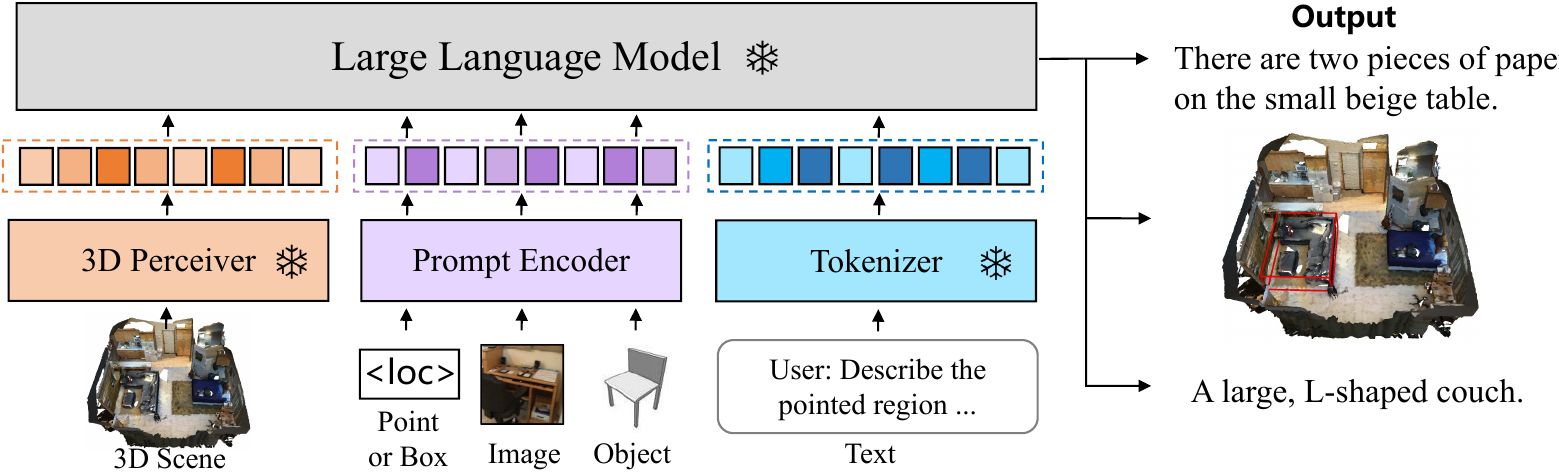}
	\caption{
        \textbf{Overview of our baseline model}. We utilize scene perceiver to extract scene tokens from 3D visual input. Multi-modal instructions are transformed into corresponding instruction tokens via their respective encoders. The scene tokens and multi-modal instruction tokens are then concatenated and fed into a frozen LLM, which generates the corresponding responses subsequently. During the training process, only the projectors are updated.
    }
	\label{fig:pipeline}
\end{figure*}

\subsection{Dataset Statistics and Analysis}
\label{dataset analysis}
\hspace{1em}\cref{tab:comparasion_dataset} presents the statistics of M3DBench. 
M3DBench contains over 320K pairs of instruction-following data. 
Among these pairs, more than 138K instructions include the interleaved multi-modal prompts we proposed.

To assess the diversity of generated instructions, we analyze the distribution of instructions based on the first word, as shown in \cref{fig:statistics} (a). 
Specifically, we extract the first word of each instruction and collected instructions starting with that word. 
Then we parse the instructions using the Natural Language Toolkit~\cite{nltk/bird2006nltk}, performing processes like tokenization and part-of-speech tagging to extract nouns and verbs from instructions. 
The findings indicate that instructions in M3DBench are diverse, including various types such as ``What" (query), ``Can" (request), ``Is" (confirmation), ``I" (first-person), ``Where" (location), and so on. 
Analyzing the word cloud of responses, as depicted in \cref{fig:statistics} (b), we observe answers pertaining to shape, color, count, action, object category, spatial relations, and so on. 
Furthermore, we demonstrated diversity in the lengths of instructions and responses, as illustrated in \cref{fig:statistics} (c) and \cref{fig:statistics} (d).


\section{Multi-modal Instruction Tuning}
\label{sec:framework}

\hspace{1em}
We introduce a baseline model that connects scenes with interleaved multi-modal instructions and accomplishes diverse tasks using a unified decoder. 
As shown in \cref{fig:pipeline}, the framework consists of three parts: scene perceiver, multi-modal instruction encoder, and LLM. 
First, the 3D scene is processed by the scene perceiver, and the features are then projected into the same feature space as the language embedding using a trainable projection layer (\cref{sec:scene tokenizer}). 
Simultaneously, prompts from different modalities within  instructions are encoded using their corresponding prompt encoder (\cref{sec:instruction tokenizer}). 
Then the visual and instruction tokens are concatenated and fed into the LLM (\cref{sec:llm decoder}).
Next, we will provide a detailed description of each module.


\subsection{3D Scene Perceiver}
\label{sec:scene tokenizer}
\hspace{1em}
Given the point cloud of a scene, denoted as $P$, we employ a pre-trained 3D encoder to extract 3D feature: 
\begin{equation}\label{equ:projection}
    f_{s} = \mathcal{E}^{3D}(P). 
\end{equation}
Similar to LLAVA~\cite{llava/liu2023visual}, we alao utilize a trainable visual feature projection matrix $\mathcal{W}^{3D}$ to project the visual features into the language embedding space and obtain scene tokens: 
\begin{equation}\label{equ:projection}
    X_{s} = \mathcal{W}^{3D} \cdot f_{s}.
\end{equation}
The scene embeddings are represented as $X_s=\left\{x_{s}^{n}\right\}_{n=1}^N$, where $x_{s}^{n}\in \mathbb{R}^{d}$ and $N$ represents the number of visual tokens. $d$ represents the dimension of hidden states in LLM.  

\subsection{Multi-modal Instruction Encoder}
\label{sec:instruction tokenizer}

\hspace{1em}
There are a total of six types of prompt formats (\cref{tab:comparasion_dataset}) in the interleaved multi-modal instructions: text, numerical coordinate, user click (point), pointed region (box), image, and 3D object.
We treat numerical \textbf{coordinate} as a specific \textbf{text}~\cite{Shikra/chen2023shikra,chatspot/zhao2023chatspot} and use the tokenizer and word embedding from LLM to obtain corresponding tokens. 
For \textbf{user click} and \textbf{pointed region}, we utilize two learnable projection matrices to extract point-level and box-level tokens, respectively. 
In the case of \textbf{image} prompt, we employ the frozen CLIP~\cite{clip/radford2021learning} to extract image features, followed by a pre-trained projector from LLaVa~\cite{llava/liu2023visual} to compute image tokens. 
For \textbf{3D object} input, we downsample them to 1024 points and normalize their coordinates into a unit sphere~\cite{scanscribe/zhu20233d}. 
Then a pre-trained encoder is used to extract object's features, and a Feed Forward Network (FFN) is inserted between the encoder and LLM to adjust these tokens.

\subsection{LLM Decoder}
\label{sec:llm decoder}

\hspace{1em} 
We utilize the pre-trained LLM~\cite{opt/zhang2022opt, llama2/touvron2023llama, vicuna/chiang2023vicuna} as a unified decoder for various vision-centric tasks. 
To accomplish this, we employ a 3D scene perceiver (\cref{sec:scene tokenizer}) to encode the input scene P into discrete scene tokens $X_s=\left\{x_{s}^{n}\right\}_{n=1}^N$.
These tokens are then concatenated with the multi-modal instruction tokens $X_i=\left\{x_{i}^{n}\right\}_{n=1}^M$. 
LLM takes both the scene tokens and the multi-modal instruction tokens as input and predicts the probability distribution of the output token $X_o=\left\{x_{o}^{n}\right\}_{n=1}^L$ in an auto-regressive manner:

\begin{equation}
    P_\theta(X_o | X_s, X_i) =  \prod_{n} P_\theta(x_o^l | x_o^{<l}; X_s, X_i).
\end{equation}
Furthermore, for tasks that rely on coordinates for assessment, such as visual grounding, we decouple them from the output of LLMs (detailed in the supplements). 
This simple approach enables us to develop a unified framework for a wide range of 3D-only tasks without the need for modifications to the existing LLMs~\cite{opt/zhang2022opt,llama/touvron2023llama,gpt-3/brown2020language}.


\begin{table*}[htbp]
    \centering
    \resizebox{\linewidth}{!}{
    \begin{tabular}{ccccccccccc}
    \toprule
      Task & 3D Vision Encoder & LLM Decoder & BLEU-1$\uparrow$ &BLEU-2$\uparrow$& BLEU-3$\uparrow$& BLEU-4$\uparrow$& ROUGE$\uparrow$ &METEOR$\uparrow$& CIDEr$\uparrow$                  \\ \hline
    \multirow{6}{*}{Dense Caption}&\multirow{3}{*}{Pointnet++~\cite{pointnet++/qi2017pointnet++}} &OPT-6.7B~\cite{opt/zhang2022opt} & 3.56& 1.43& 0.52& 0.21& 14.18& 9.79& 17.01\\
                                    & &LLaMA-2-7B~\cite{llama2/touvron2023llama} &10.60 &\textbf{4.53} &\textbf{1.70} &\textbf{0.73} &\textbf{18.70} &\textbf{13.40} &22.05 \\
                                    & &Vicuna-7B-v1.5~\cite{vicuna/chiang2023vicuna} &2.97 & 1.04& 0.32& 0.00& 11.78& 9.04&13.88 \\                \cline{2-3}
    &\multirow{3}{*}{Transformer~\cite{transformer/vaswani2017attention}} &OPT-6.7B~\cite{opt/zhang2022opt} & 10.72& 4.44& 1.45& 0.0& 14.58& 10.35&\textbf{23.76} \\ 
                                   & &LLaMA-2-7B~\cite{llama2/touvron2023llama} & 10.07&3.71 &1.38 & 0.0&17.32 &12.03 &20.72 \\
                                   & &Vicuna-7B-v1.5~\cite{vicuna/chiang2023vicuna} & \textbf{11.96}& 4.38& 1.28& 0.0& 14.13& 9.46& 23.72\\ \hline
    \multirow{6}{*}{Visual Question Answering}&\multirow{3}{*}{Pointnet++~\cite{pointnet++/qi2017pointnet++}} &OPT-6.7B~\cite{opt/zhang2022opt} &57.45 & 49.48& 43.57& 38.78& 58.34& 30.30& 336.96\\
                                    & &LLaMA-2-7B~\cite{llama2/touvron2023llama} &\textbf{61.01} & \textbf{53.35}& \textbf{47.63}& \textbf{43.00}& \textbf{61.59}& \textbf{32.05}& \textbf{379.05}\\
                                    & &Vicuna-7B-v1.5~\cite{vicuna/chiang2023vicuna} &46.30 &38.13 & 32.20& 27.56&51.55 & 27.03 &239.98 \\                \cline{2-3}
    &\multirow{3}{*}{Transformer~\cite{transformer/vaswani2017attention}} &OPT-6.7B~\cite{opt/zhang2022opt} &57.26 &50.35 & 44.97& 40.50& 59.55& 30.64& 365.60\\ 
                                   & &LLaMA-2-7B~\cite{llama2/touvron2023llama} & 60.23&52.41 &47.02 &42.61 & 59.24& 30.96& 356.42\\
                                   & &Vicuna-7B-v1.5~\cite{vicuna/chiang2023vicuna} & 17.77& 14.22& 11.86& 10.07& 22.12& 11.32& 95.98\\ \hline
    \multirow{6}{*}{Embodied Question Answering}&\multirow{3}{*}{Pointnet++~\cite{pointnet++/qi2017pointnet++}} &OPT-6.7B~\cite{opt/zhang2022opt} &\textbf{47.55} &37.69 &30.91 &24.44 &49.17 &\textbf{26.04} &212.12 \\
                                    & &LLaMA-2-7B~\cite{llama2/touvron2023llama} &45.85 &35.92 &29.32 &22.79 &48.34 &24.89 &194.09 \\
                                    & &Vicuna-7B-v1.5~\cite{vicuna/chiang2023vicuna} & 21.09& 15.61& 12.28& 9.41& 44.06& 20.55& 169.72\\                \cline{2-3}
    &\multirow{3}{*}{Transformer~\cite{transformer/vaswani2017attention}} &OPT-6.7B~\cite{opt/zhang2022opt} & 47.37&\textbf{37.86} &\textbf{31.33} &\textbf{24.76} &\textbf{50.83} &25.95 &\textbf{218.01} \\ 
                                   & &LLaMA-2-7B~\cite{llama2/touvron2023llama} & 44.20& 33.86&27.49 &21.58 & 45.83& 22.74& 179.33\\
                                   & &Vicuna-7B-v1.5~\cite{vicuna/chiang2023vicuna} &38.24& 29.71& 24.63& 19.64& 40.62&21.00&155.12  \\ \hline

    \multirow{6}{*}{Multi-region Reasoning}&\multirow{3}{*}{Pointnet++~\cite{pointnet++/qi2017pointnet++}} &OPT-6.7B~\cite{opt/zhang2022opt} & \textbf{57.53}& \textbf{50.03}& \textbf{43.57}& 38.27& 61.23& 33.74& 363.87\\
                                & &LLaMA-2-7B~\cite{llama2/touvron2023llama} &56.24 &49.32 &43.42 &\textbf{38.46} &\textbf{61.48} &\textbf{34.01} &\textbf{378.17} \\
                                & &Vicuna-7B-v1.5~\cite{vicuna/chiang2023vicuna} &47.98 &39.18 &32.28 &26.82 &49.87 &27.59 &212.93 \\                \cline{2-3}
    &\multirow{3}{*}{Transformer~\cite{transformer/vaswani2017attention}} &OPT-6.7B~\cite{opt/zhang2022opt} &36.92 &30.78 &25.91 &21.60 &44.51 &24.27 &240.89 \\ 
                                   & &LLaMA-2-7B~\cite{llama2/touvron2023llama} & 55.00&47.88 &42.31 & 37.60& 59.90&32.56 &351.96 \\
                                   & &Vicuna-7B-v1.5~\cite{vicuna/chiang2023vicuna} & 21.96& 17.21& 13.87& 11.06&27.07 &12.68 &95.40 \\ \hline
    \multirow{6}{*}{Eobodied Planning}&\multirow{3}{*}{Pointnet++~\cite{pointnet++/qi2017pointnet++}} &OPT-6.7B~\cite{opt/zhang2022opt} &49.22 &41.11 &35.04 &29.71 &50.90 &26.65 &133.94 \\
                                & &LLaMA-2-7B~\cite{llama2/touvron2023llama} & 57.66& 50.18& 44.86& 40.76& 56.46& 29.77& \textbf{253.09}\\
                                & &Vicuna-7B-v1.5~\cite{vicuna/chiang2023vicuna} &21.68 &15.27 &10.87 &8.10 &32.73 &19.78 &83.39 \\                \cline{2-3}
    &\multirow{3}{*}{Transformer~\cite{transformer/vaswani2017attention}} &OPT-6.7B~\cite{opt/zhang2022opt} & \textbf{59.47}& \textbf{53.24}& \textbf{48.08}& \textbf{43.46}& \textbf{61.14}&\textbf{33.34} &213.15 \\ 
                                   & &LLaMA-2-7B~\cite{llama2/touvron2023llama} & 52.98&45.17 &39.05 &34.27 &49.95 & 28.70& 171.51\\
                                   & &Vicuna-7B-v1.5~\cite{vicuna/chiang2023vicuna} & 37.50& 30.71& 25.33& 20.54& 38.55& 21.50& 114.91\\ 
    \bottomrule
    \end{tabular}
    }
    \caption{
        \textbf{Benchmark for multiple tasks: Dense Caption (DC), Visual Question Answering (VQA), Embodied Question Answering (EQA), Multi-region Reasoning (MR), Embodied Planning (EP).} We present the performance of baseline methods on our evaluation dataset. ↑ means the higher, the better.
    }
    \vspace{8pt}
    \label{tab:benchmark-5tasks}
\end{table*}

\subsection{Training Strategy}
\hspace{1em} 
The training objective is to maximize the likelihood of generating this target response sequence $X_o=\left\{x_{o}^{n} \right\}_{n=1}^L$, given the visual input $X_s$ and multi-modal instruction $X_i$:

\begin{equation}
    \mathcal{L_\theta} = -\sum_{n=1}^{L} \log P_\theta (x_o^l | x_o^{<l}; X_s, X_i).
\end{equation}
Here, $\theta$ represents the trainable parameters. 
%
Note that during training, we freeze the 3D encoder, image encoder, as well as language decoder, and only train all the projection layers to enable rapid iterations. 
Exploring alternative architecture or refining the training strategy could potentially yield further improvements. 
We leave this as a direction for future work.

\section{Experiments}
\label{sec:benchmark}

\hspace{1em} 
We first introduce the baseline model, metrics, and implementation details in~\cref{implementation}.
Additionally, we provide a benchmark on 3D scene understanding, reasoning and description in~\cref{quantitative}.
Finally, we showcase some visualization results in ~\cref{sec:qualitative}.
More details, quantitative results, and qualitative examples are provided in supplements.

\subsection{Baseline, Metrics, and Implementations}
\label{implementation}

\myparagraph{Baseline.} 
Since no prior method that works out of the box with our interleaved multi-modal instruction setup, we develop several variant models as baseline based on LLM~\cite{opt/zhang2022opt,llama2/touvron2023llama,vicuna/chiang2023vicuna} to accommodate M3DBench. 
Specifically, we incorporate two different types of 3D encoders, based on PointNet++~\cite{pointnet++/qi2017pointnet++} and Transformer~\cite{transformer/vaswani2017attention}, into our baseline model. Furthermore, we consider three versions of LLMs as our language decoder: OPT-6.7B~\cite{opt/zhang2022opt}, LLaMA-2-7B~\cite{llama2/touvron2023llama}, and Vicuna-7B-v1.5~\cite{vicuna/chiang2023vicuna}. 
After end-to-end instruction tuning, we evaluate baseline models on the evaluation dataset to assess their effectiveness.

\myparagraph{Evaluation Metrics.}
The evaluation metrics include both traditional and GPT metrics. 
Traditional metrics, such as CiDEr~\cite{cider/vedantam2015cider}, METEOR~\cite{meteor/banerjee2005meteor}, Acc@0.25IoU~\cite{scanrefer/chen2020scanrefer}, and so on, are used to measure the model's performance on specific tasks. 
For a more comprehensive evaluation of the models' instruction-following abilities, we employ GPT-4 to assess the quality of the different variants' responses. 
Specifically, we provide GPT-4 with the answers generated by different variant models, the reference answers, and evaluation requirements. 
GPT-4 evaluates these responses and assigns a score ranging from 0 to 100. 
A higher average score indicates better performance of the model. 
Furthermore, we request GPT-4 to provide justifications for the scoring results, which helps us better judge the validity of the evaluation.

\myparagraph{Implementations.}
Following previous works in 3D learning~\cite{dc_vote2cap/chen2023end, 3detr/misra2021end}, we downsample each 3D scene to 40,000 points as our scene input. 
For the PointNet++-based 3D encoder, we initialize it with the checkpoint obtained from Depth Contrast~\cite{depthcontrast/zhang2021self}. 
As for the Transformer-based encoder, we employ the checkpoint from Vote2Cap-DETR~\cite{dc_vote2cap/chen2023end}. 
Additionally, we use the pre-trained encoder ViT-L/14~\cite{clip/radford2021learning} as our image feature encoder. 
We train all the baseline models using the Adam optimizer~\cite{adamw/loshchilov2017decoupled} with a cosine annealing scheduler where the learning rate decays from $10^{-5}$ to $10^{-6}$.
%
Our batch size is set to 2 during training, utilizing 4 Nvidia A100 (40G) GPUs, which allows us to complete the training within 2 days.

\subsection{Quantitative Evaluation}
\label{quantitative}

\myparagraph{Understanding,  Reasoning, and Planning.} 
To establish a benchmark for scene understanding, reasoning, and planning, we comprehensively evaluated six variant models and reported the quantitative results on our evaluation dataset. 
\cref{tab:benchmark-5tasks} presents the performance of baselines across five tasks: Dense Captioning (DC), Visual Question Answering (VQA), Embodied Question Answering (EQA), Multi-region Reasoning (MR), and Embodied Planning (EP). 
We employed BLEU 1-4~\cite{bleu/papineni2002bleu}, ROUGE-L~\cite{rouge/lin2004rouge}, METEOR~\cite{meteor/banerjee2005meteor}, and CiDEr~\cite{cider/vedantam2015cider} as evaluation metrics.

Analyzing the results, one can see that when using the same language decoder, the Pointnet++~\cite{pointnet++/qi2017pointnet++}-based models underperformed compared to the Transformer~\cite{instruction-tuning/ouyang2022training}-based models in the DC and EP tasks, while outperformed them in the MR task.
However, upon switching the language decoder while keeping the 3D encoder constant, Vicuna-7B-v1.5~\cite{vicuna/chiang2023vicuna} exhibited lower overall performance compared to other LLMs across almost all tasks.
The evaluation of our benchmark dataset suggests a diversity in the performance of MLMs, with each demonstrating unique strengths and weaknesses across diverse tasks.
Moreover, the suboptimal performance of current baseline models across various tasks offers potential direction for further development of 3D MLMs. 
For instance, enhancing the performance of MLMs on benchmark tasks such as scene understanding, perception, and planning is crucial and we leave them for future work to explore.



\begin{table}[htbp]
    \vspace{12pt}
    \centering
    \resizebox{0.9\linewidth}{!}{
    \begin{tabular}{ccc}
    \toprule
    3D Vision Encoder &  LLM Decoder & GPT-4 Score                 \\ \hline
    \multirow{3}{*}{Pointnet++~\cite{pointnet++/qi2017pointnet++}}      &OPT-6.7B~\cite{opt/zhang2022opt} & 9.87  \\
                                                  &LLaMA-2-7B~\cite{llama2/touvron2023llama} &27.89   \\
                                                  &Vicuna-7B-v1.5~\cite{vicuna/chiang2023vicuna} & \textbf{32.37}  \\ \cline{1-2}
    \multirow{3}{*}{Transformer~\cite{transformer/vaswani2017attention}}             &OPT-6.7B~\cite{opt/zhang2022opt} &  16.84 \\
                                                  &LLaMA-2-7B~\cite{llama2/touvron2023llama} & 27.37  \\
                                                  &Vicuna-7B-v1.5~\cite{vicuna/chiang2023vicuna} &  29.08 \\
    \bottomrule
    \end{tabular}
    }
    \caption{
        \textbf{Benchmark for detailed description.} In practice, we randomly select 39 scenes and provide detailed descriptions generated by GPT-4 for each scene. We then assess the relative scores achieved by different variants. Responses generated by variant models and GPT-4' descriptions are fed back into GPT-4 for comparative analysis and scoring, along with the provision of relevant explanations for each variant's answer. Experiments demonstrate that the model based on Vicuna-7B-V1.5~\cite{vicuna/chiang2023vicuna} demonstrated superior performance.
    }
    \vspace{4pt}
    \label{tab:benchmark-dc}
\end{table}

\myparagraph{Detailed Description.}
As shown in \cref{tab:benchmark-dc}, regarding detailed description capabilities, utilizing PointNet++~\cite{pointnet++/qi2017pointnet++} as the scene encoder and Vicuna-7B-V1.5~\cite{vicuna/chiang2023vicuna} as the language decoder yielded the best performance, surpassing the sub-optimal variant by a margin of $+3.29$ points. 
%
Another observation is that all variants based on OPT~\cite{opt/zhang2022opt} demonstrated relatively lower performance. 
Furthermore, we note that overall, all baseline models demonstrate inferior performance, suggesting that current baseline models possess limited capabilities in handling detailed descriptions. 
In supplements, we provide a qualitative presentation of the description results and the criteria for GPT-4 scoring.

\subsection{Qualitative Results}
\label{sec:qualitative}

\begin{figure}[t]
	\centering
	\includegraphics[width=\linewidth]{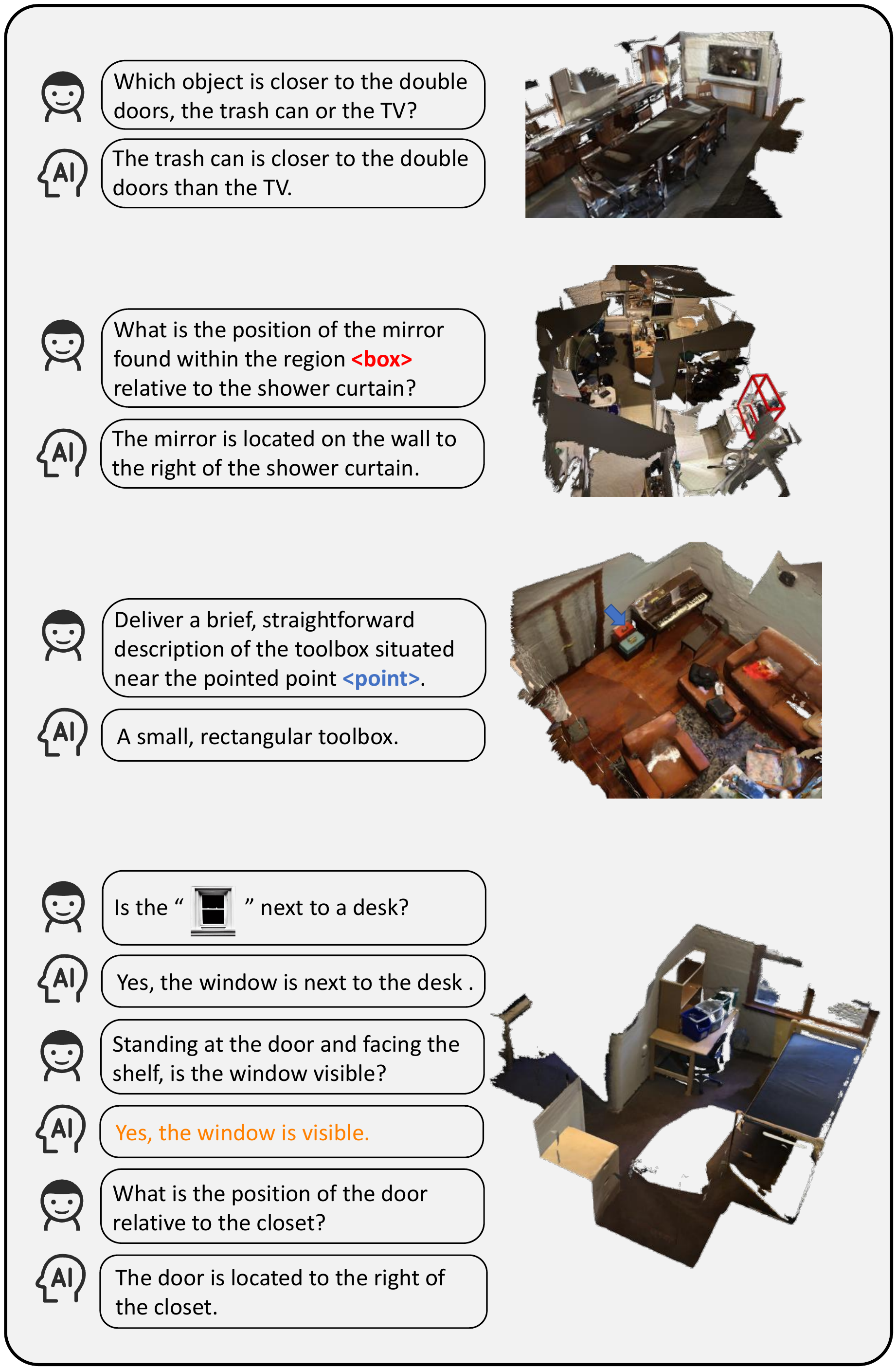}
	\caption{
    	\textbf{Qualitative Results.}  We provide visualization results on various 3D-centric tasks in diverse 3D environments. \orange{Orange} highlights the wrong answer.
    }
	\label{fig:vis}
\end{figure}

\hspace{1em} 
We showcase some qualitative examples of our baseline model on the evaluation dataset in \cref{fig:vis}. One can see that our proposed method, trained on M3DBench, is capable of performing corresponding tasks under a variety of interleaved multi-modal instructions.

\section{Conclusion}
\label{sec:discussion}
\hspace{1em} 
In this paper, we present M3DBench, a comprehensive multi-modal 3D instruction-following dataset, designed to facilitate the development of MLMs in the 3D domain. 
M3DBench encompasses a wide range of 3D vision-centric tasks and over 320K pairs of 3D instruction-following pairs, covering fundamental functionalities such as visual perception, scene understanding, spatial reasoning, planning, and navigation.
Additionally, M3DBench introduces a novel multi-modal prompting scheme, interweaving language instruction with coordinate, image, pointed region, and other visual prompts. 
We also develop a simple yet efficient baseline model to validate the effectiveness of M3DBench, providing benchmarks for multiple tasks. 
Comprehensive quantitative and qualitative results demonstrate that models trained with M3DBench can successfully follow human instructions and complete 3D visual-related tasks. 
We hope that our proposed multi-modal 3D instruction dataset, baseline model, and benchmarks will inspire and fuel future explorations in the field of 3D MLMs.

{\small
\bibliographystyle{ieee_fullname}
\bibliography{reference}
}

\clearpage
\onecolumn
\section*{
    {\hfill \LARGE Supplementary Material \hfill }
}
\vspace{80pt}
\renewcommand\thesection{\Alph{section}}
\setcounter{section}{0}
\noindent This supplementary material provides further details about M3DBench (\cref{app:dataset}), quantitative experiments (\cref{app:benchmark}) on multi-round dialogue and 3D localization, additional experiments for held-out evaluation (\cref{app:held-out}), implementation details (\cref{app:implementation}) of baseline model and prompt for GPT-4 evaluation (\cref{app:gpt-4}). 



\section{Dataset}
\label{app:dataset}

In \cref{app:dataset:example}, we provide more examples in M3Dbench for each task. Following that, we will introduce the dataset construction in ~\cref{app:dataset:construction} and provide statistics for the evaluation dataset in ~\cref{app:dataset:statistics}.

\subsection{More Examples}
\label{app:dataset:example}

\begin{figure*}[htbp]
	\centering
	\includegraphics[width=0.85\linewidth]{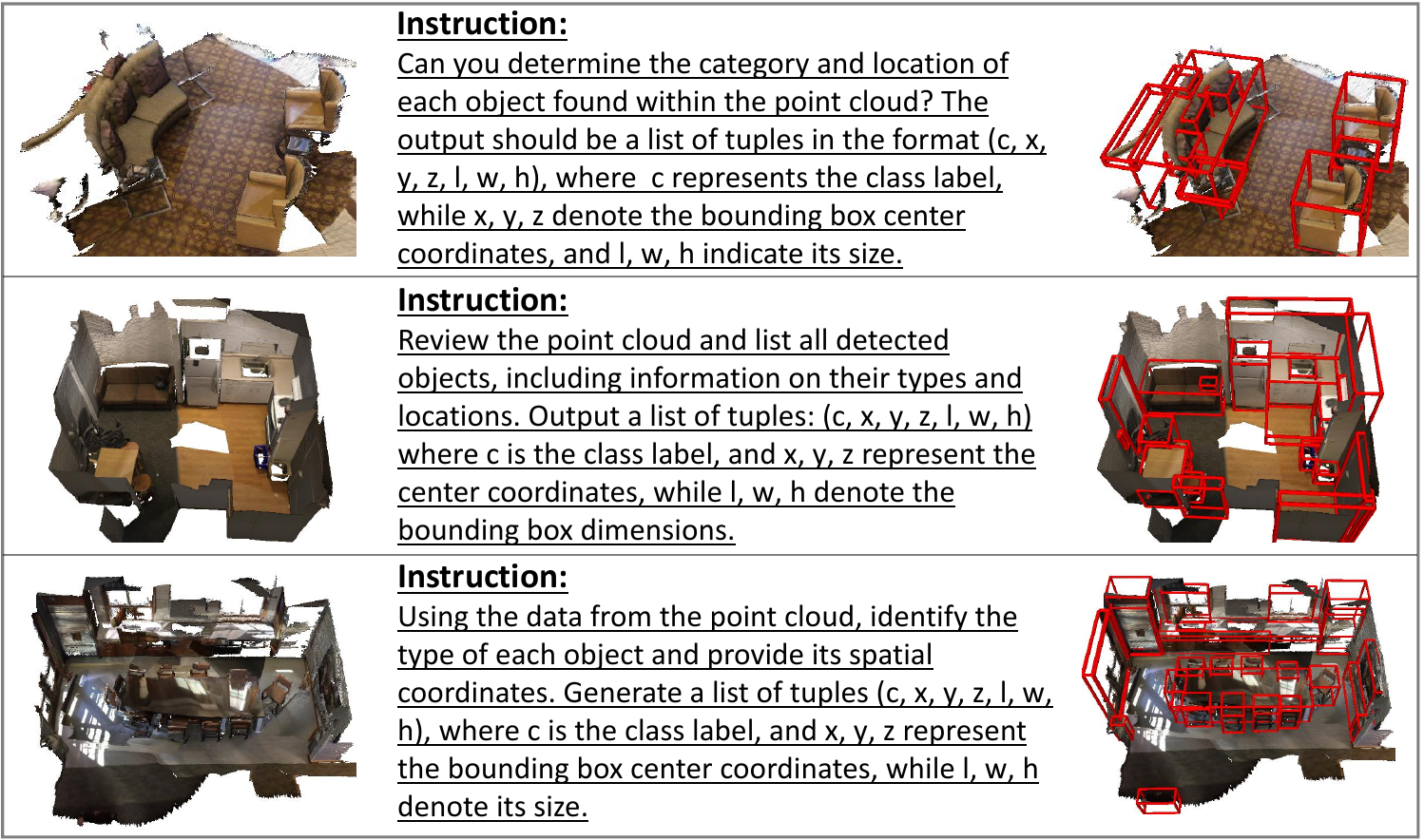}
	\caption{
        \textbf{Examples of 3D object detection.} The left column represents the 3D scene, the middle column displays the instructions, and the right column shows the annotations for the object detection task. We save annotations in textual format and for visualization purposes here, we extract the bounding boxes from the text.
        }
	\label{fig:od}
    \vspace{80pt}
\end{figure*}

\begin{figure*}[htbp]
	\centering
	\includegraphics[width=0.85\linewidth]{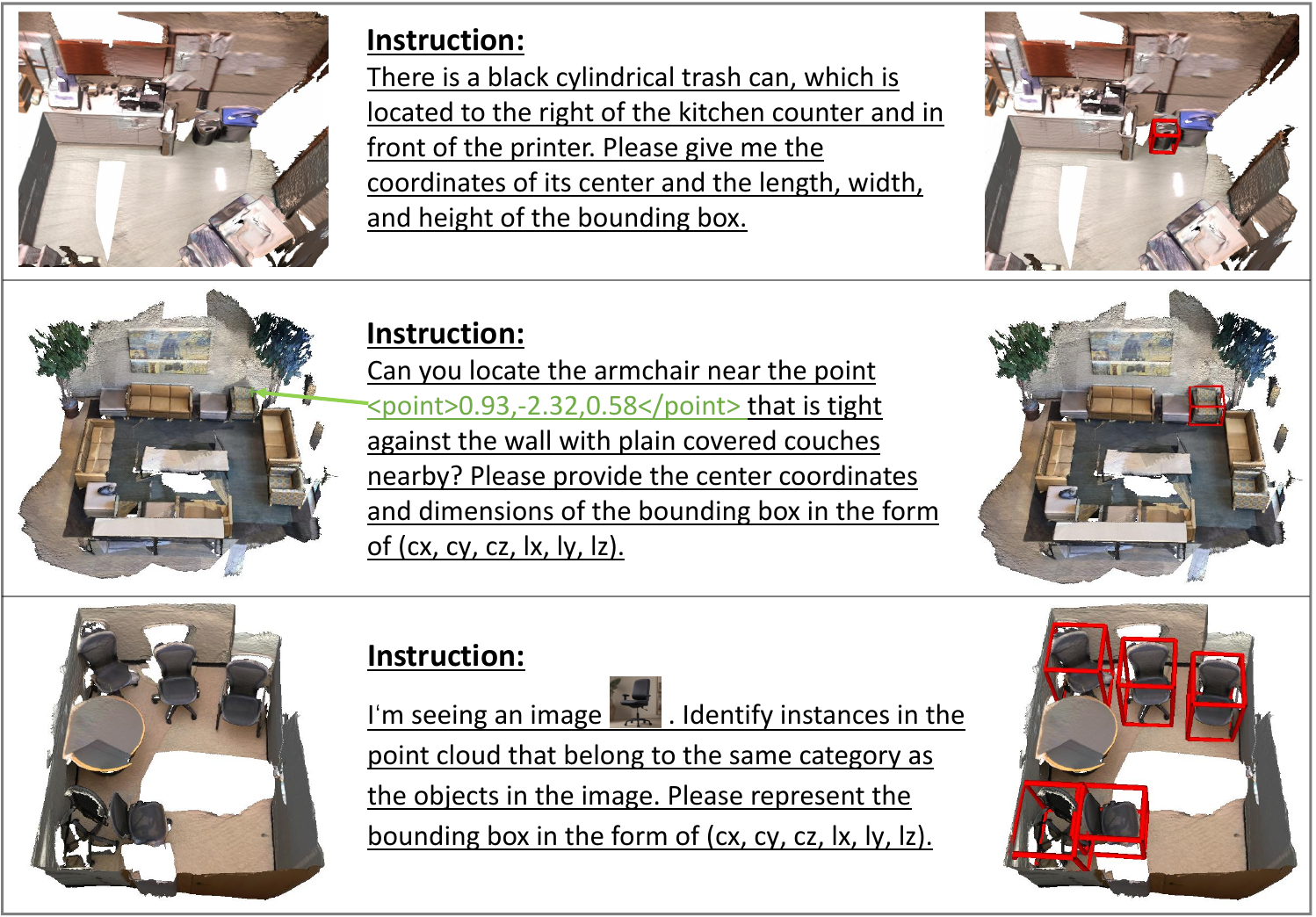}
	\caption{
        \textbf{Examples of 3D visual grounding.} The left column represents the 3D scene, the middle column displays the instructions, and the right column shows the annotations for the visual grounding. M3DBench includes interleaved multi-modal instructions, and the annotations extend beyond annotating a single target object, encompassing the identification of multiple objects.
        }
	\label{fig:vg}
    \vspace{5pt}
\end{figure*}

\begin{figure*}[htbp]
	\centering
	\includegraphics[width=0.85\linewidth]{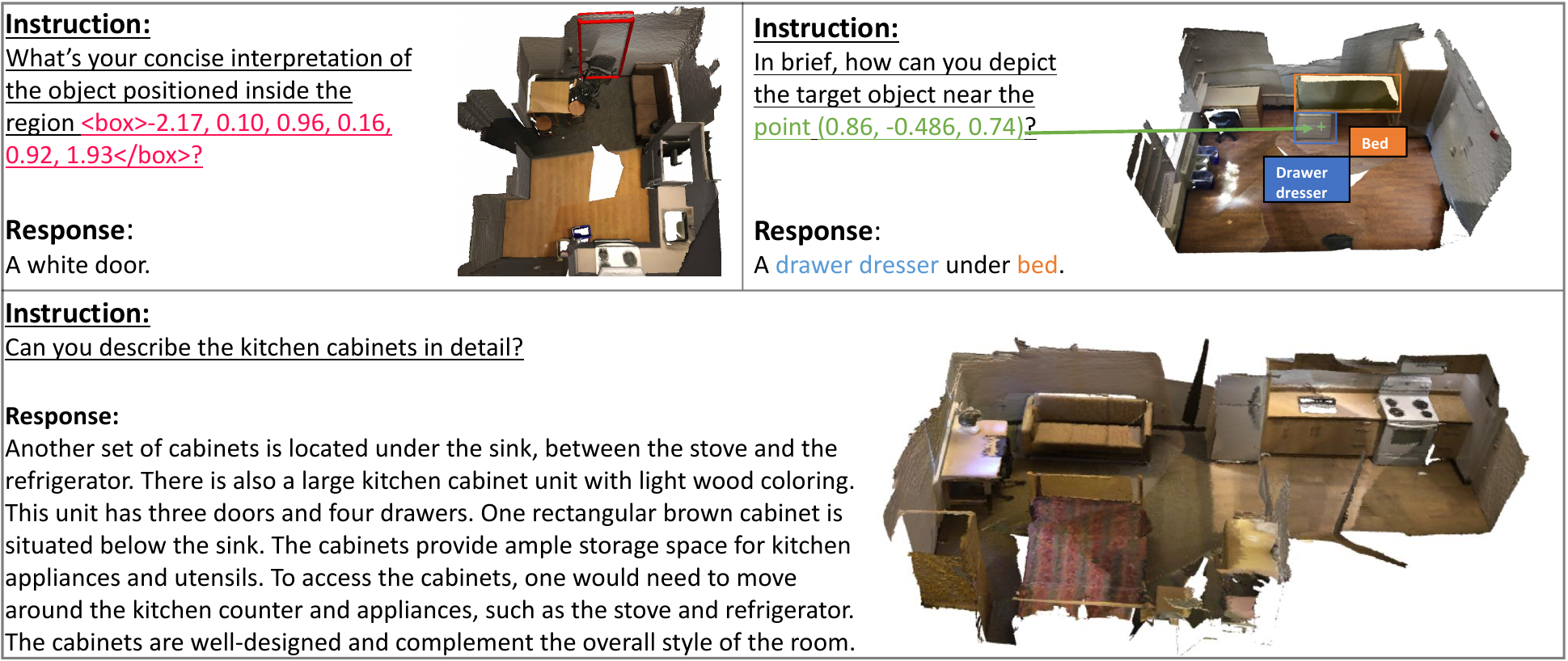}
	\caption{
        \textbf{Examples of 3D dense caption.} We design diverse multi-modal instructions for dense captions for M3DBench. Additionally, we introduce terms such as \textit{brief} or \textit{detailed} within instructions to generate either concise titles or detailed descriptions for objects.
        }
	\label{fig:dc}
    \vspace{25pt}
\end{figure*}

\begin{figure*}[htbp]
	\centering
	\includegraphics[width=0.85\linewidth]{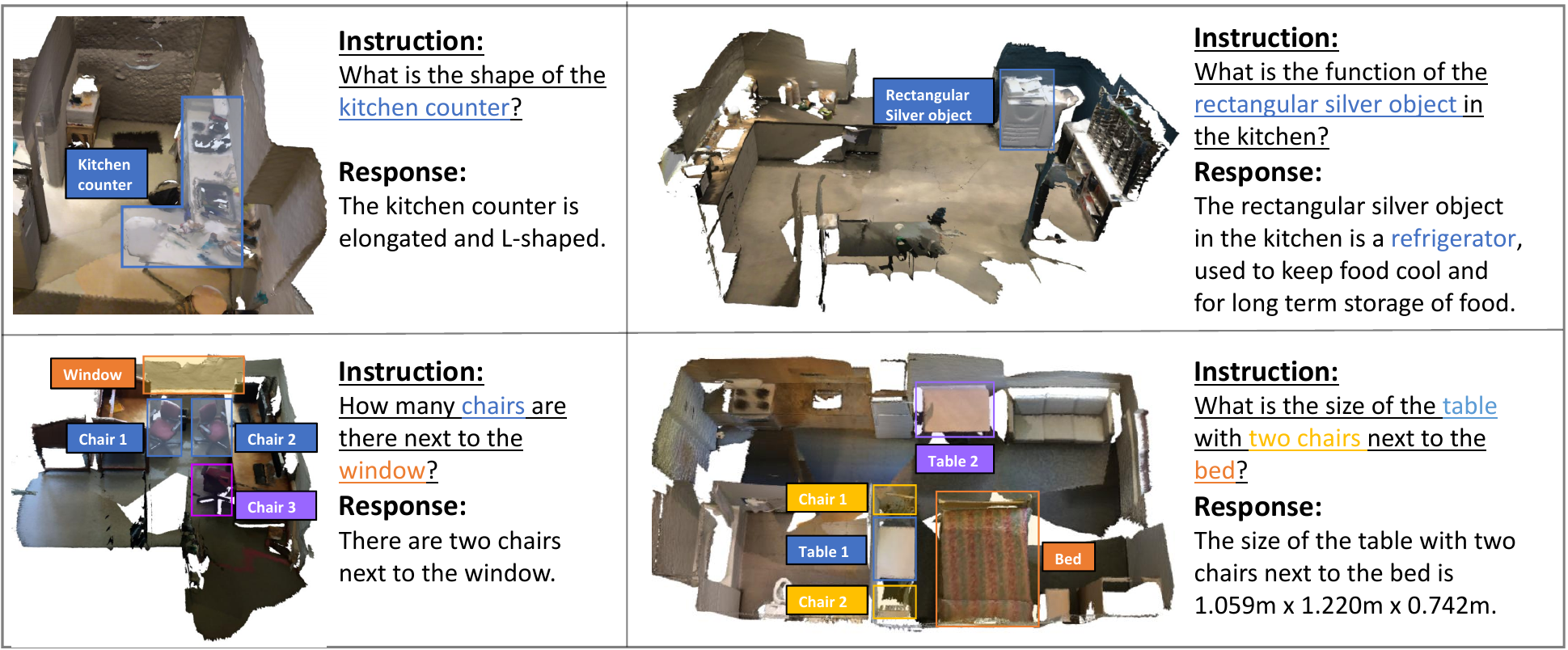}
	\caption{
        \textbf{Examples of 3D visual question answering.} M3DBench comprises open-ended, free-form questions involving instance location, shape and size, object count, scene type, object and room functionality, and more. For instance, when asked about the functionality of a \textit{rectangular silver object} in the upper-right scene, the answer begins by identifying the object and then describing its functionality. Furthermore, the two examples below illustrate instances where there might be multiple objects of \textit{the same category} in the scene.}
	\label{fig:vqa}
    \vspace{5pt}
\end{figure*}

\begin{figure*}[htbp]
	\centering
	\includegraphics[width=0.85\linewidth]{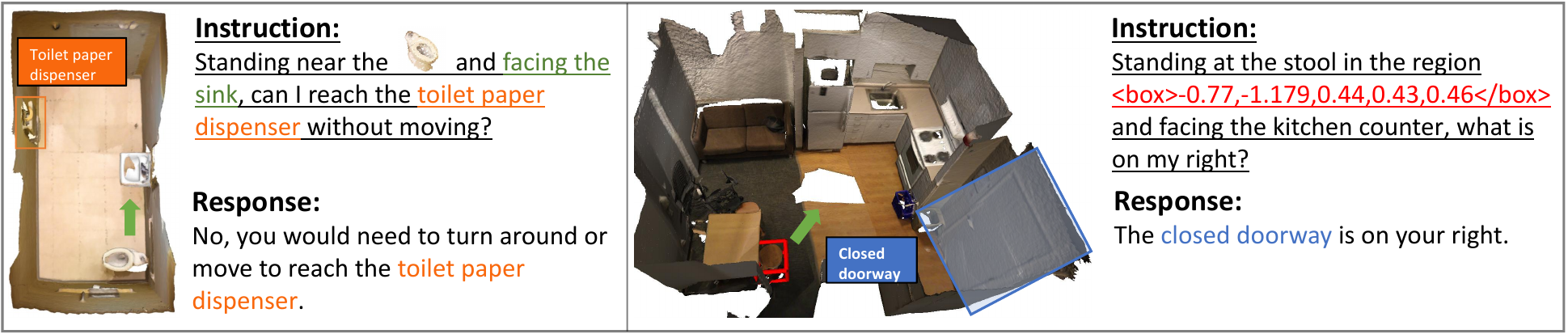}
	\caption{
        \textbf{Examples of embodied question answering.} Embodied question answering requires the agent to understand the surrounding environment in order to answer questions under that situation.
        }
	\label{fig:eqa}
    \vspace{5pt}
\end{figure*}

\begin{figure*}[htbp]
	\centering
	\includegraphics[width=0.85\linewidth]{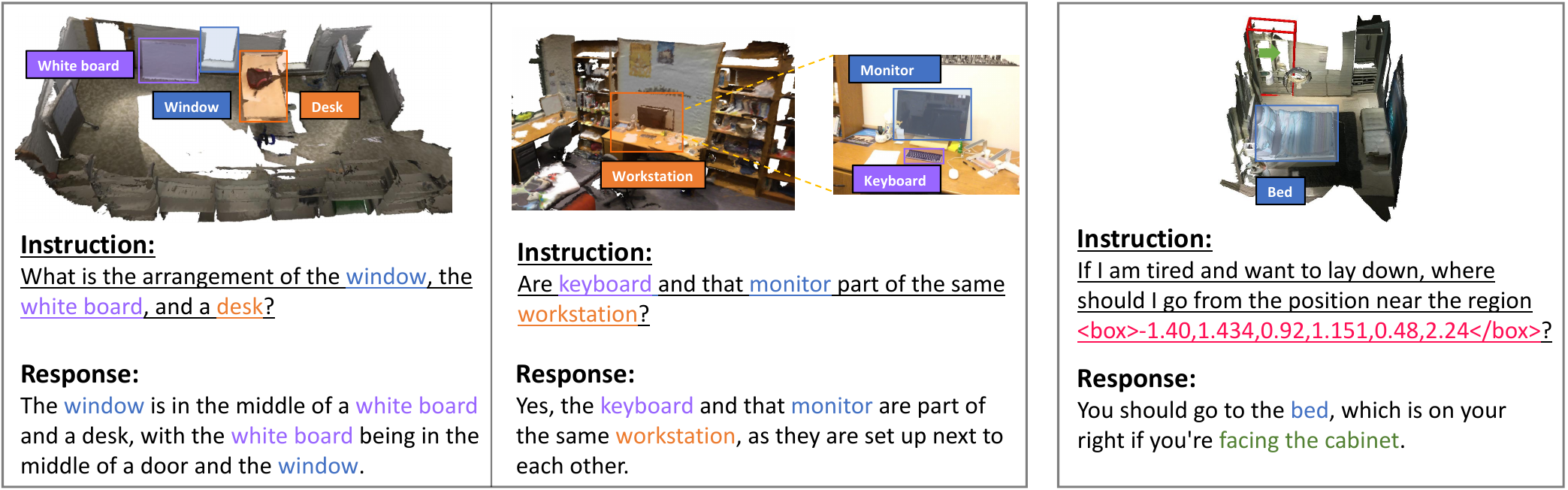}
	\caption{
        \textbf{Examples of multi-region reasoning (left) and embodied planning (right).} In multi-region reasoning tasks (left), at least two objects are involved, querying their relative relationships and sizes, which enables a detailed comprehension of the scene. On the other hand, embodied planning (right) requires an agent to perceive the environment, understand the user’s intentions, and then generate appropriate responses or actions to achieve predetermined goals.
        }
	\label{fig:mr_ep}
        \vspace{40pt}
\end{figure*}

\begin{figure*}[htbp]
	\centering
	\includegraphics[width=0.85\linewidth]{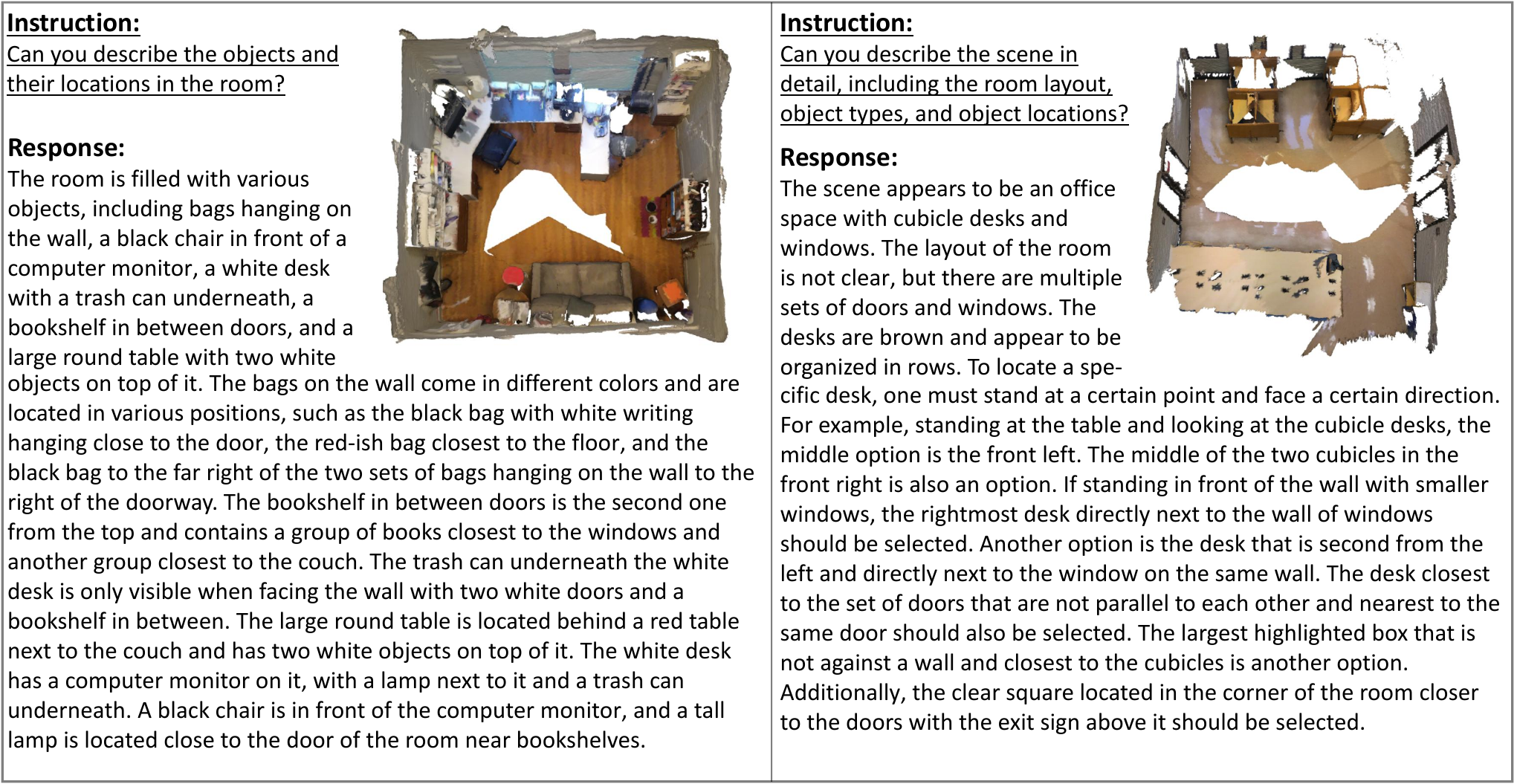}
	\caption{
        \textbf{Examples of scene description.} Scene description requires integrating the visual information of the entire scene to generate a detailed and comprehensive introduction. This description will encompass aspects such as objects within the scene, the type of room, its functionality, and so on.
        }
	\label{fig:sd}
    \vspace{5pt}
\end{figure*}

\begin{figure*}[htbp]
	\centering
	\includegraphics[width=0.85\linewidth]{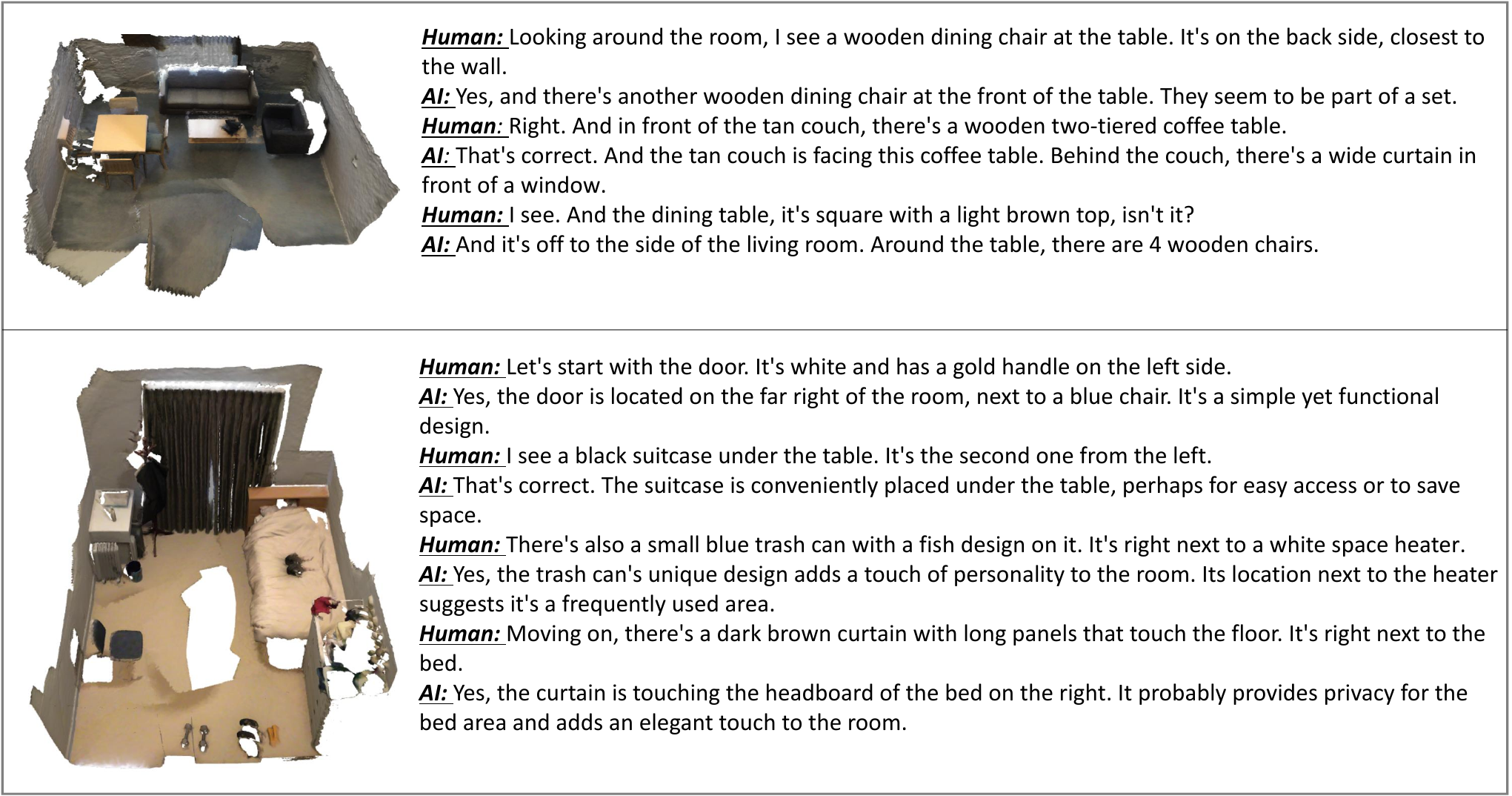}
	\caption{
        \textbf{Examples of multi-round dialogue.} Multi-round dialogue necessitates the agent's ability to engage in natural and coherent communication with humans. This capability involves not only understanding and generating language but also ensuring accuracy and coherence in context.
        }
	\label{fig:md}
    \vspace{5pt}
\end{figure*}

\begin{figure*}[htbp]
	\centering
	\includegraphics[width=0.85\linewidth]{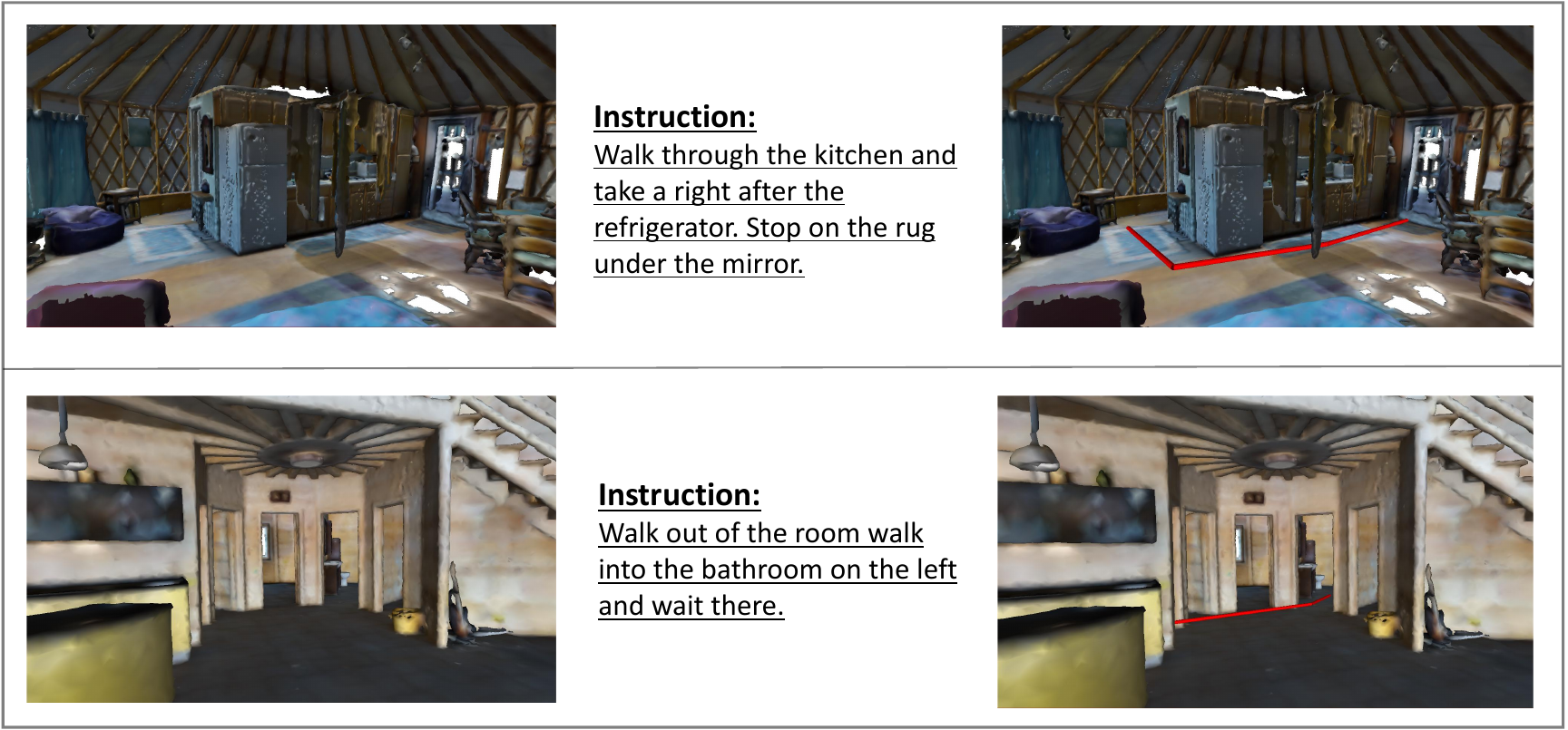}
	\caption{
        \textbf{Examples of 3D vision language navigation.} The 3D scene is depicted in the left column, instructions are presented in the middle column, and annotations for the vision language navigation task are shown in the right column. Annotations are stored in textual format, and for visual representation here, we extract the pathway from the text.
        }
	\label{fig:vln}
    \vspace{5pt}
\end{figure*}

\subsection{Dataset Construction}
\label{app:dataset:construction}
In this work, we introduce a comprehensive 3D instruction tuning dataset, M3DBench, which serves as the foundation for developing versatile and practical general-purpose assistants in the real-world 3D environment. 
M3DBench comprises 3D data from publicly available datasets~\cite{scannet/dai2017scannet,mp3d/chang2017matterport3d,referit3d/achlioptas2020referit3d,scanrefer/chen2020scanrefer,phraserefer/yuan2022toward,nav/krantz2020beyond, shapenet/chang2015shapenet}, along with interleaved multi-modal instructions and responses generated using self-instruct methods\cite{self-instruct/wang2022self} and GPTs~\cite{gpt-3/brown2020language,gpt-4/openai2023gpt}. 
From \cref{table:prompt_dc,table:prompt_qa,table:prompt_eqa,table:prompt_mr,table:prompt_sd,table:prompt_md,table:prompt_ep}, we provide detailed description of the prompts designed for various 3D tasks, each comprising system messages and manually crafted context examples. 
For tasks such as object detection, we manually design instruction and response templates, then replace the template's keywords with annotations to construct instruction-response data~\cite{lamm/yin2023lamm}, as illustrated in \cref{table:template_od} and \cref{table:template_dc}.
Furthermore, we have developed an interleaved multi-modal instruction formula by substituting corresponding templates for the \textit{\textless{target}\textgreater} in the instructions, as shown in \cref{table:multimodal_prompt}.

\subsection{Evaluation Dataset}
\label{app:dataset:statistics}
\begin{figure*}[htbp]
	\centering
	\includegraphics[width=0.7\linewidth]{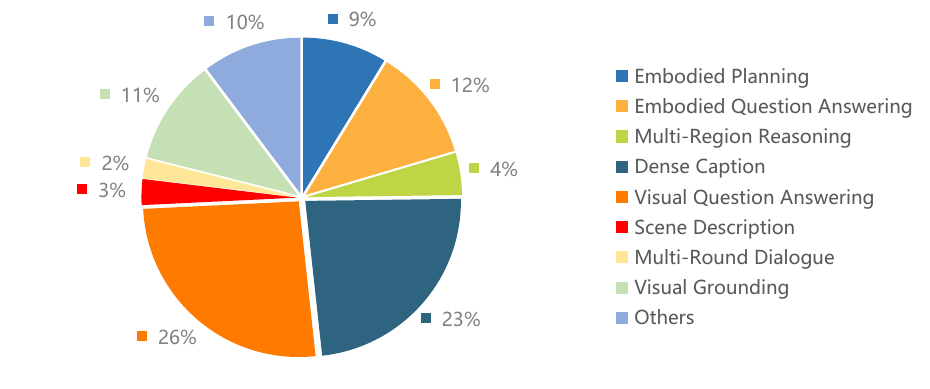}
	\caption{
        \textbf{The evaluation dataset covers a range of fundamental abilities within real-world 3D environments,} such as visual perception, scene comprehension, spatial reasoning, and embodied planning.
        }
	\label{fig:app_sta}
    \vspace{5pt}
\end{figure*}

To quantitatively evaluate the effectiveness of instruction-tuned MLMs, we constructed an evaluation dataset to assess the models' performance across various dimensions such as visual perception, scene understanding, spatial reasoning, and embodied planning. Our evaluation dataset consists of over 1.5K data samples, distributed as shown in \cref{fig:app_sta}.

\clearpage
\begin{table*}[htbp]
	\centering
	\includegraphics[width=0.9\linewidth]{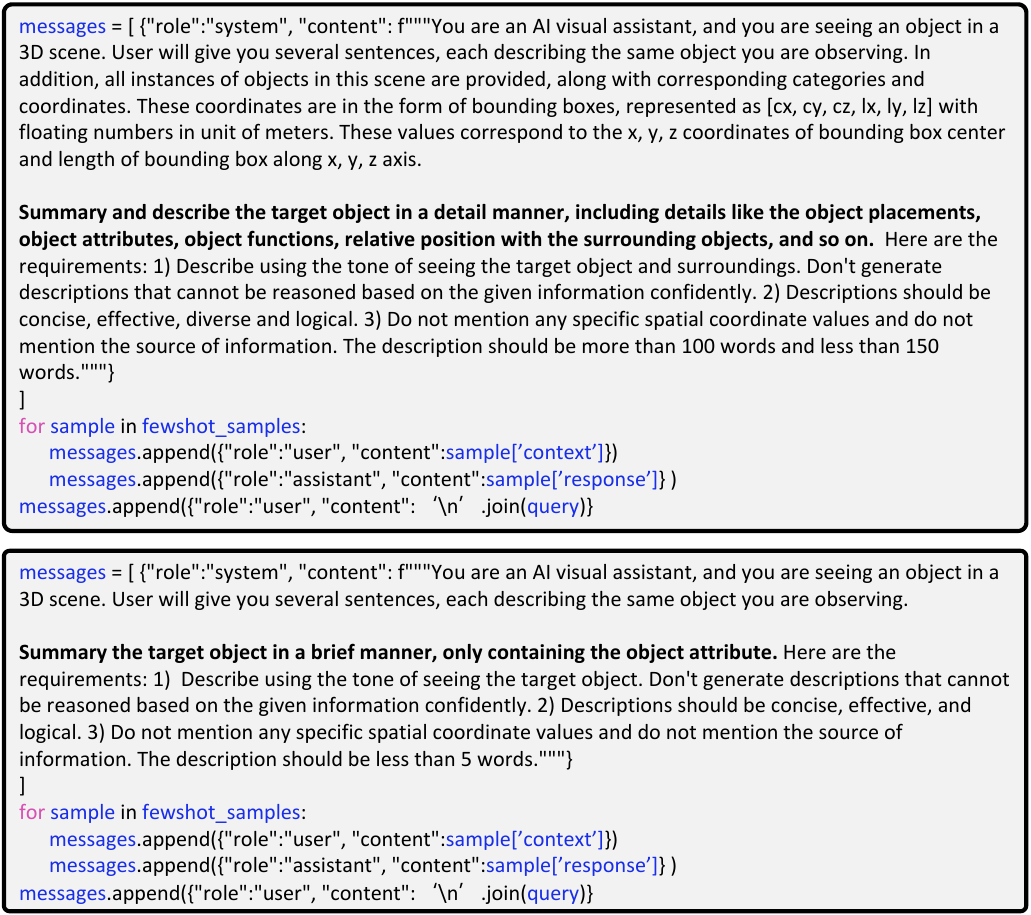}
	\caption{
        System message used to generate detailed (top) and brief (bottom) dense caption data in M3DBench.
        }
	\label{table:prompt_dc}
    \vspace{50pt}
\end{table*}

\begin{table*}[htbp]
	\centering
	\includegraphics[width=0.9\linewidth]{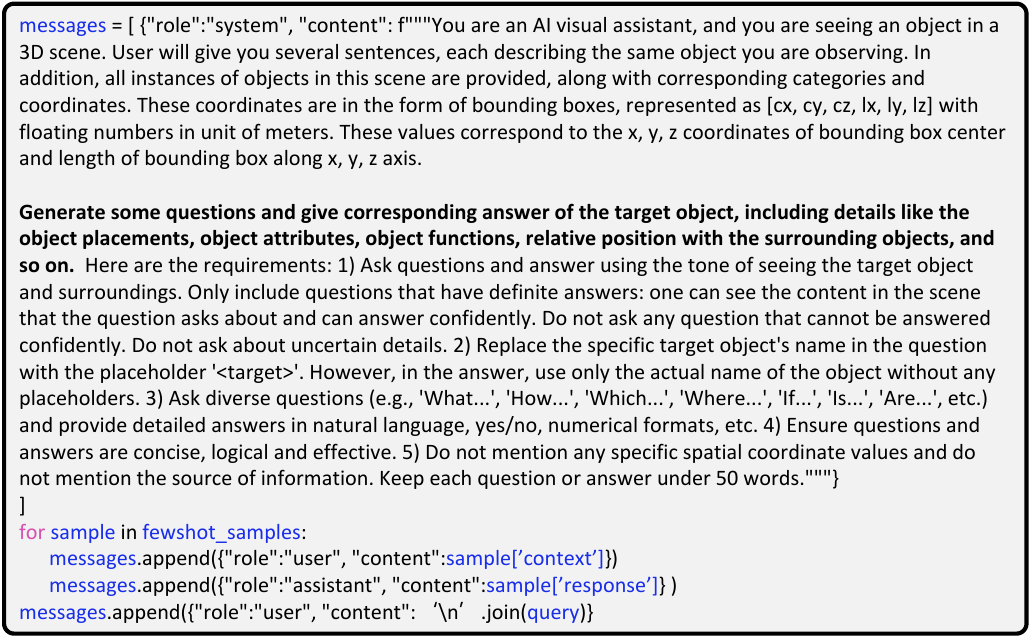}
	\caption{
        System message used to generate instruction-response pairs for visual question answering in M3DBench.
        }
	\label{table:prompt_qa}
    \vspace{5pt}
\end{table*}

\begin{table*}[htbp]
	\centering
	\includegraphics[width=0.9\linewidth]{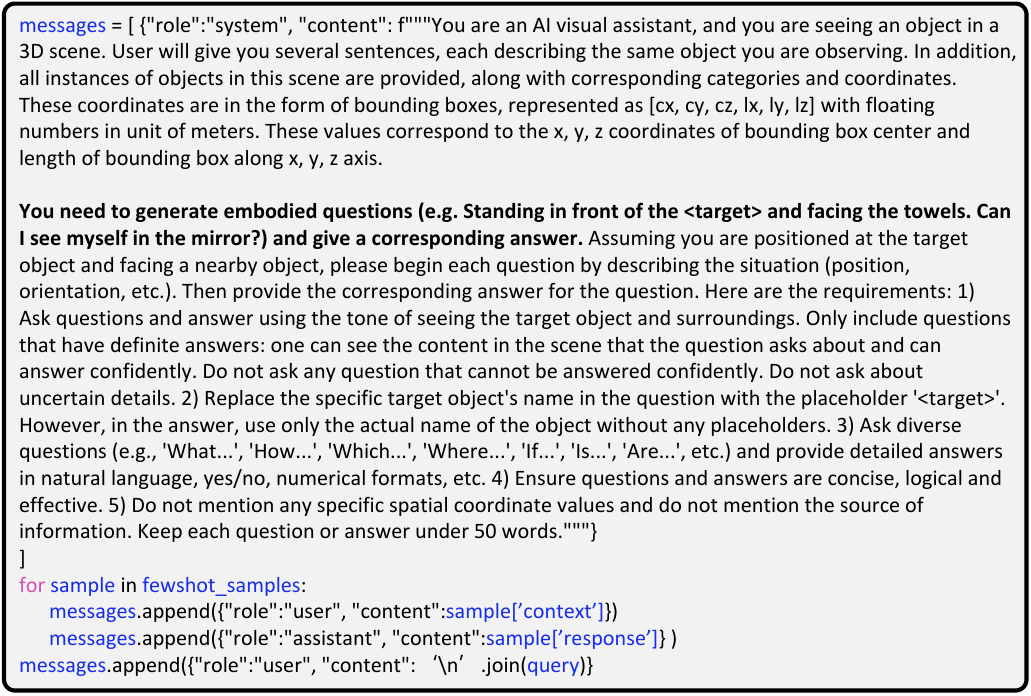}
	\caption{
        System message used to generate instruction-response pairs for embodied question answering in M3DBench.
        }
	\label{table:prompt_eqa}
    \vspace{5pt}
\end{table*}

\begin{table*}[htbp]
	\centering
	\includegraphics[width=0.9\linewidth]{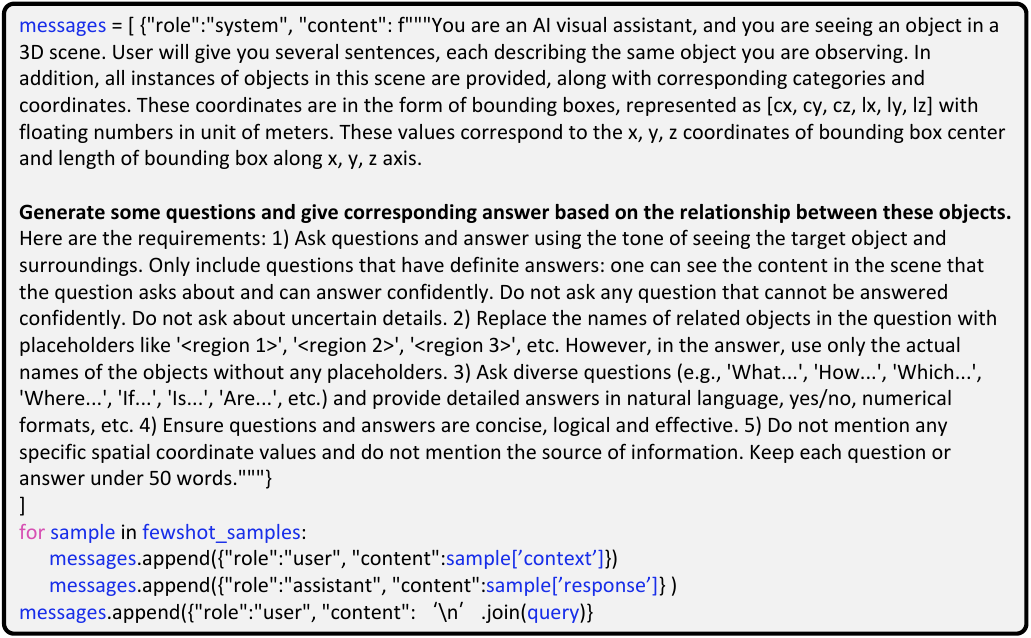}
	\caption{
        System message used to generate instruction-response pairs for multi-region reasoning in M3DBench.
        }
	\label{table:prompt_mr}
    \vspace{5pt}
\end{table*}

\begin{table*}[htbp]
	\centering
	\includegraphics[width=0.9\linewidth]{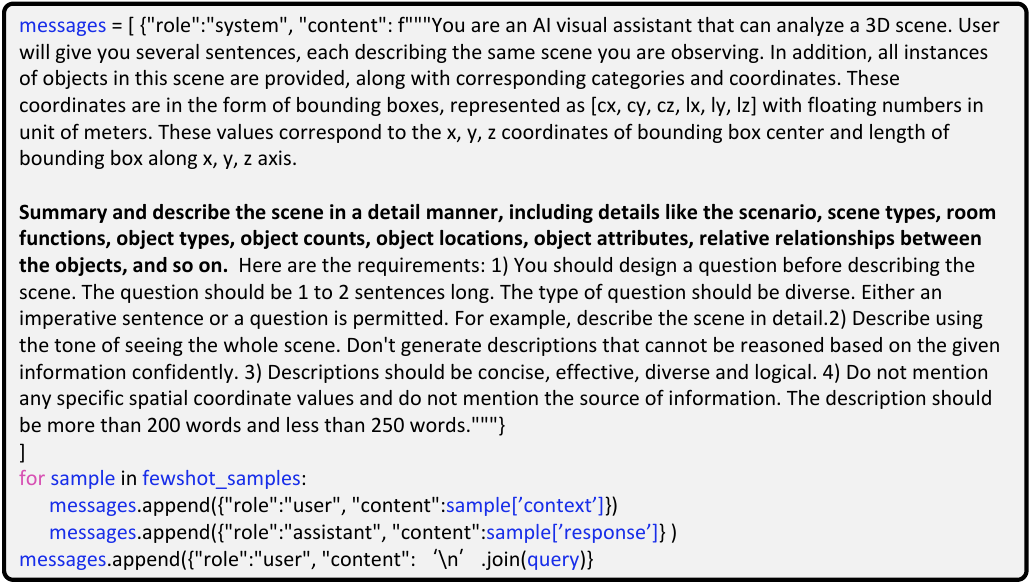}
	\caption{
        System message used to generate instruction-response pairs for scene description in M3DBench.
        }
	\label{table:prompt_sd}
    \vspace{5pt}
\end{table*}

\begin{table*}[htbp]
	\centering
	\includegraphics[width=0.9\linewidth]{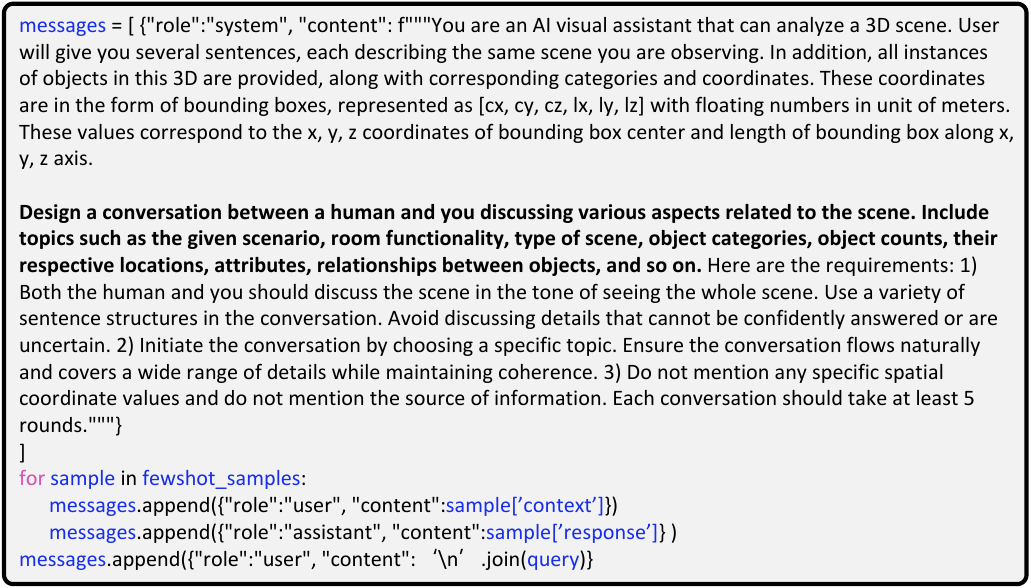}
	\caption{
        System message used to generate instruction-response pairs for multi-round dialogue in M3DBench.
        }
	\label{table:prompt_md}
    \vspace{5pt}
\end{table*}

\begin{table*}[htbp]
	\centering
	\includegraphics[width=0.9\linewidth]{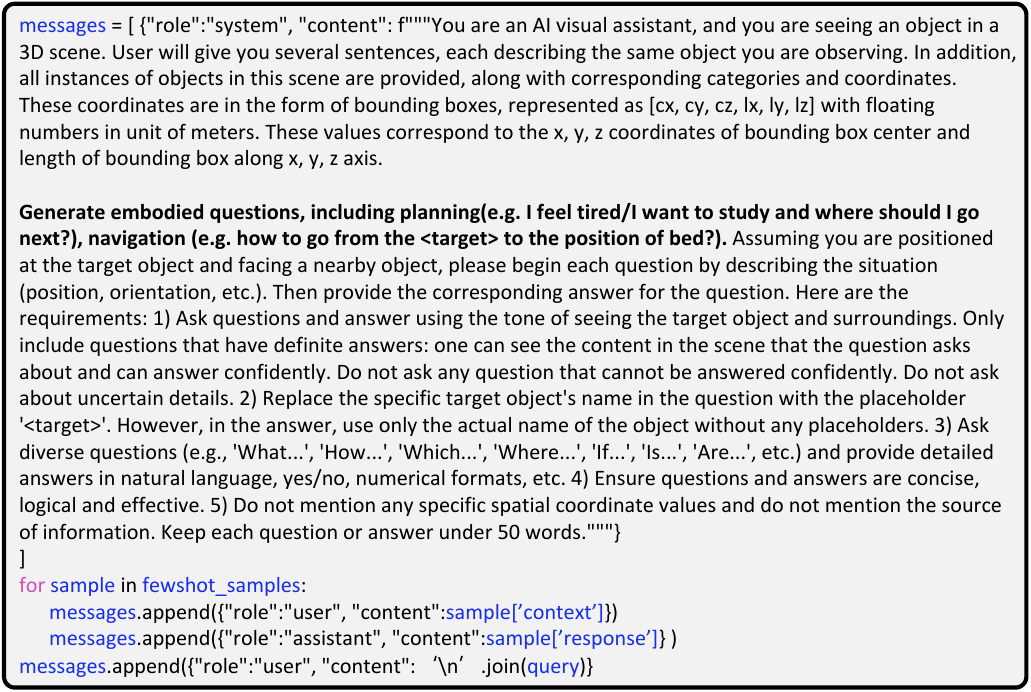}
	\caption{
        System message used to generate instruction-response pairs for embodied planning in M3DBench.
        }
	\label{table:prompt_ep}
    \vspace{5pt}
\end{table*}

\clearpage
\begin{table*}[htbp]
	\centering
	\includegraphics[width=0.9\linewidth]{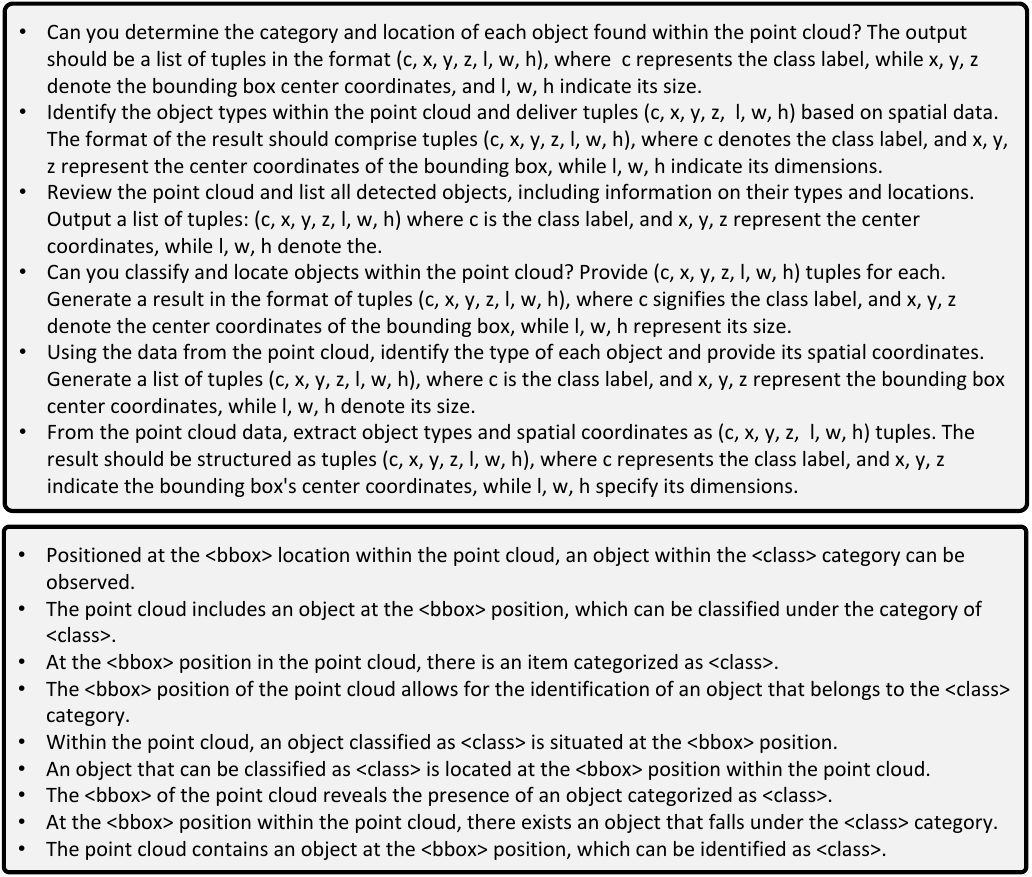}
	\caption{
        Some examples of question (top) and answer (bottom) templates for 3D object detection in M3DBench.
        }
	\label{table:template_od}
    \vspace{5pt}
\end{table*}

\clearpage
\begin{table*}[htbp]
	\centering
	\includegraphics[width=0.9\linewidth]{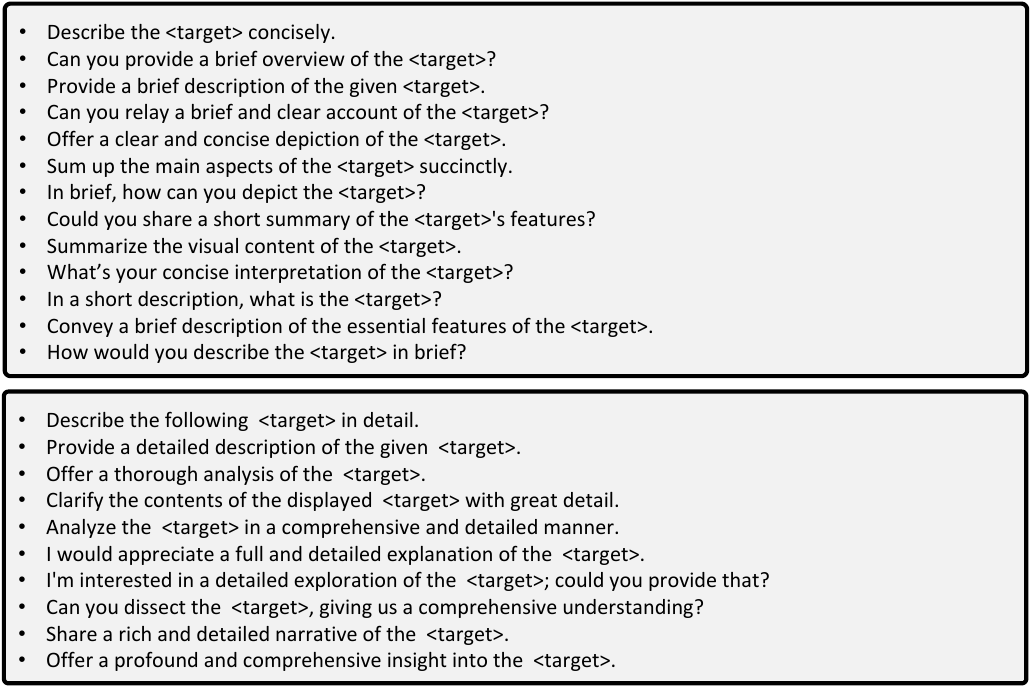}
	\caption{
        Some examples of instructions for brief (top) and detailed (bottom) dense caption in M3DBench.
        }
	\label{table:template_dc}
    \vspace{5pt}
\end{table*}

\clearpage
\begin{table*}[t]
	\centering
	\includegraphics[width=0.9\linewidth]{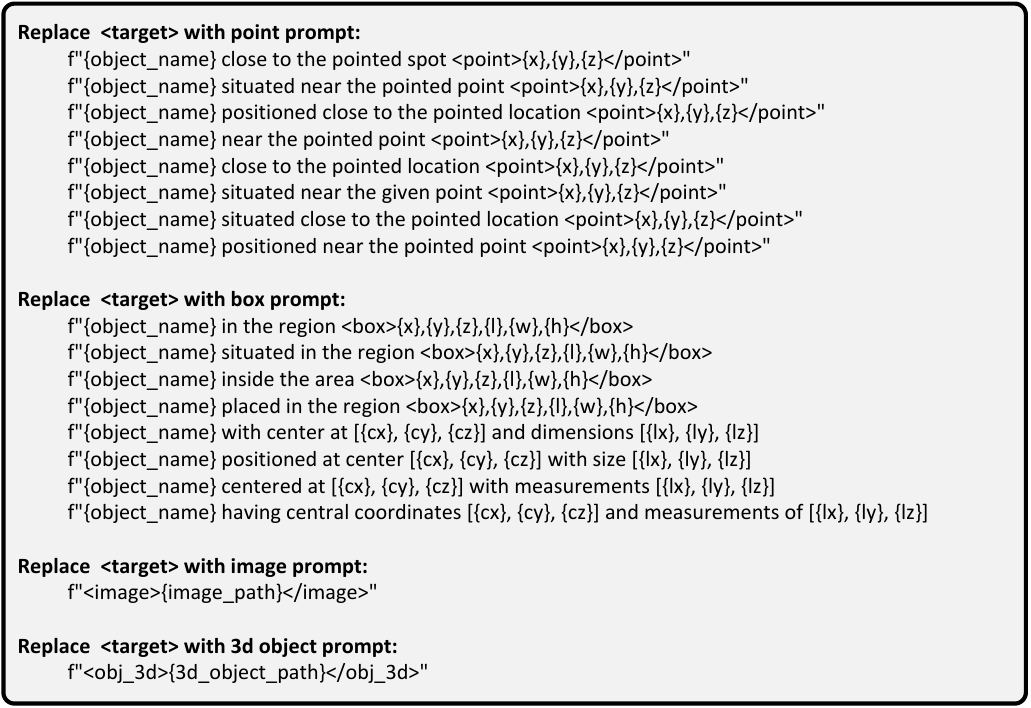}
	\caption{
        The formula for interleaved multi-modal instruction generation in M3DBench.
        }
	\label{table:multimodal_prompt}
\end{table*}

\section{Experiments on Dialogue and Localization} 
\label{app:benchmark}

\myparagraph{Quantitative Evaluation on Multi-round Dialogue.}
We score the conversational abilities of the baseline models using GPT-4~\cite{gpt-4/openai2023gpt}. 
For multi-round dialogue, the baseline model employing PointNet++~\cite{pointnet++/qi2017pointnet++} as the scene encoder and Vicuna-7B-V1.5~\cite{vicuna/chiang2023vicuna} as the language decoder demonstrates the optimal performance, surpassing the next best variant by $+0.71$ points. 
Similar to the conclusions derived from the results of detailed description (detailed in the \cref{quantitative}), all OPT-based variants~\cite{opt/zhang2022opt} exhibit relatively lower performance. 
In \cref{app:gpt-4}, we provide prompts used for scoring conversations with GPT-4~\cite{gpt-4/openai2023gpt}, along with qualitative results for multi-turn dialogues and the GPT-4~\cite{gpt-4/openai2023gpt} scoring criteria.

\begin{table}[htbp]
    \centering
    \vspace{10pt}
    \begin{minipage}[htbp]{0.45\linewidth}
        \flushleft
        \makeatletter\def\@captype{table}
        \resizebox{\linewidth}{!}{
        \begin{tabular}{ccc}
        \toprule
        3D Vision Encoder &  LLM Decoder & Relative Score                 \\ \hline
        \multirow{3}{*}{Pointnet++~\cite{pointnet++/qi2017pointnet++}}      &OPT-6.7B~\cite{opt/zhang2022opt} & 40.97  \\
                                                      &LLaMA-2-7B~\cite{llama2/touvron2023llama} & 44.74  \\
                                                      &Vicuna-7B-v1.5~\cite{vicuna/chiang2023vicuna} & \textbf{46.06}  \\ \cline{1-2}
        \multirow{3}{*}{Transformer~\cite{transformer/vaswani2017attention}}             &OPT-6.7B~\cite{opt/zhang2022opt} &  29.52 \\
                                                      &LLaMA-2-7B~\cite{llama2/touvron2023llama} & 38.61  \\
                                                      &Vicuna-7B-v1.5~\cite{vicuna/chiang2023vicuna} &  45.35 \\
        \bottomrule
        \end{tabular}
        }
        \caption{
            \textbf{Benchmark for multi-round dialogue.} \textit{Relative Score} is generated by the GPT-4~\cite{gpt-4/openai2023gpt}, based on the evaluation of the model's response.
        }
        \label{table:app:dialogue}
    \end{minipage}
    \hspace{0.005\linewidth}
    \begin{minipage}[htbp]{0.45\linewidth}
        \flushright
        \makeatletter\def\@captype{table}
        \resizebox{\linewidth}{!}{
        \begin{tabular}{ccc}
        \toprule
        3D Vision Encoder &  LLM Decoder & Acc$@0.25IoU$                 \\ \hline
        \multirow{2}{*}{Pointnet++~\cite{pointnet++/qi2017pointnet++}}      &OPT-6.7B~\cite{opt/zhang2022opt} & 3.09 \\
                                                      &LLaMA-2-7B~\cite{llama2/touvron2023llama} &  1.60\\ \cline{1-2}
        \multirow{2}{*}{Transformer~\cite{transformer/vaswani2017attention}}             &OPT-6.7B~\cite{opt/zhang2022opt} & 1.22 \\
                                                      &LLaMA-2-7B~\cite{llama2/touvron2023llama} &  \textbf{3.57} \\
        \bottomrule
        \end{tabular}
        }
        \caption{
            \textbf{Benchmark for object localization.} We assess the baseline model's ability to identify and localize objects in the 3D scene. Specifically, the baseline model is tasked with outputting the location of the target object a given specific instruction. The metric utilized is Acc$@0.25IoU$.
        }
        \label{table:app:localization}
    \end{minipage}
    
\end{table}

\myparagraph{Quantitative Evaluation on 3D object Localization.}
For the 3D object localization task (i.e., finding the object in a scene that best matches a given instruction), we propose using a unified output format to represent object position. 
To acquire localization data, we derive 3D bounding boxes from the ``[$cx$, $cy$, $cz$, $l$, $w$, $h$]" provided in the generated text. 
Here, $cx$, $cy$, $cz$ correspond to the x, y, z coordinates of the bounding box center, while $l$, $w$, $h$ represent the size of the bounding box along the x, y, z axes. 
For each value defining the 3D bounding box, we retain one decimal place. 
In \cref{table:app:localization}, we present baseline performances regarding 3D localization. 
Results indicate that our proposed baseline model exhibits suboptimal performance on localizing. We leave the improvement of MLMs' abilities in 3D scene perception and localization for future work.

\section{Held-out Evaluation}
\label{app:held-out}

\myparagraph{Training and Evaluation Protocols.}
In order to assess the zero-shot performance of the  baseline model fine-tuned on the multi-modal instruction data for unseen tasks, we partition our dataset into two types: held-in and held-out datasets. 
Specifically, we consider embodied question answering (EQA) and embodied planning (EP) from M3DBench as unseen tasks, with their corresponding dataset (held-out dataset) excluded during the training process. 
We train the baseline model on the training dataset for the remaining tasks (held-in dataset) and evaluate the model's performance using the validation set from the held-out dataset.

\myparagraph{Baselines and Metrics.}
We utilize a pre-trained masked transformer encoder\cite{dc_vote2cap/chen2023end} as the scene encoder and employ two large language models, OPT-6.7B~\cite{opt/zhang2022opt} and LLaMA-2-7B~\cite{llama2/touvron2023llama}, as the decoder in our baseline model. 
Furthermore, we employ BLEU 1-4~\cite{bleu/papineni2002bleu}, ROUGE-L~\cite{rouge/lin2004rouge}, METEOR~\cite{meteor/banerjee2005meteor}, and CiDEr~\cite{cider/vedantam2015cider} as evaluation metrics.

\begin{table*}[htbp]
    \centering
    \vspace{15pt}
    \resizebox{\linewidth}{!}{
    \begin{tabular}{cccccccccc}
    \toprule
      Task & LLM Decoder & BLEU-1$\uparrow$ &BLEU-2$\uparrow$& BLEU-3$\uparrow$& BLEU-4$\uparrow$& ROUGE$\uparrow$ &METEOR$\uparrow$& CIDEr$\uparrow$                  \\ \hline
    \multirow{2}{*}{Embodied Question Answering} & OPT-6.7B\cite{opt/zhang2022opt}& 28.76  & 21.67  &17.51  & 13.96  & 30.78 &  17.64  &139.06 \\
                                     &LLaMA-2-7B~\cite{llama2/touvron2023llama}&\textbf{35.76}&\textbf{27.42}&\textbf{21.89}&\textbf{16.83}&\textbf{40.04}&\textbf{20.47}&\textbf{163.71} \\  \hline       
    
\multirow{2}{*}{Embodied Planning} &OPT-6.7B\cite{opt/zhang2022opt} & 21.13  &16.07   &12.36  &8.99  &28.96  &16.28  & 47.62 \\
                                    &LLaMA-2-7B~\cite{llama2/touvron2023llama}  & \textbf{33.80}&\textbf{25.15}&\textbf{19.23}&\textbf{14.71}&\textbf{33.30}&\textbf{19.65}&\textbf{58.21}  \\

    \bottomrule
    \end{tabular}
    }
    \caption{
        \textbf{Zero-shot results on Embodied Question Answering (EQA) and Embodied Planning (EP).} For held-out evaluation, we demonstrate the performance of baseline methods on two tasks. The upward arrow (↑) indicates that higher values represent better performance. Notably, we find that leveraging LLaMA-2~\cite{llama2/touvron2023llama} as the language decoder exhibits superior zero-shot generalization compared to the OPT-based~\cite{opt/zhang2022opt} model.
        }
    \vspace{8pt}
    \label{tab:heldout}
\end{table*}

\myparagraph{Result Analysis.}
In \cref{tab:heldout}, we present the performance of the baseline model for held-out evaluation. 
Additionally, we compare baselines using different LLMs as language decoders. 
All baselines follow the same training and evaluation protocols described above. 
In summary, we draw three insights:
1) through instruction tuning and multi-task learning on the held-in dataset of M3DBench, the baseline model exhibits reasoning ability when dealing with tasks that it hasn't encountered before.
2) LLaMA-based~\cite{llama2/touvron2023llama} model outperforms the baseline model based on OPT~\cite{opt/zhang2022opt} in zero-shot generalization.
3) There remain gaps in zero-shot results compared to results from full supervised instruction fine-tuning (detailed in the \cref{quantitative}).
These findings indicate that through instruction tuning and multi-task learning on M3DBench, our model demonstrates reasoning abilities on tasks that haven't encountered before.
This emphasizes the significance of instruction tuning for achieving zero-shot generalization.

\section{Implementation}
\label{app:implementation}

\myparagraph{Scene Encoder.}
As introduced in \cref{implementation}, we employ two commonly used types of 3D pre-trained feature extractors as scene encoders: one based on PointNet++~\cite{pointnet++/qi2017pointnet++} and the other based on Transformer~\cite{transformer/vaswani2017attention}. 
The PointNet++-based scene encoder comprises four layers for feature extraction and down-sampling, coupled with two layers for feature aggregation and up-sampling~\cite{depthcontrast/zhang2021self}. 
The final layer generates features for sampled points, from which we derive scene-level 256-dimensional features via global max-pooling. 
In addition, the Transformer-based encoder initially tokenizes input point clouds into 2048 point tokens through a set abstraction layer~\cite{pointnet++/qi2017pointnet++}, followed by three cascaded Transformer encoder blocks with masking radii of 0.16, 0.64, and 1.44~\cite{dc_vote2cap/chen2023end}. 
Between the first two Transformer blocks, there is an additional set abstraction layer that downsamples the encoded tokens, with each token represented by 256 dimensions.

\myparagraph{Multi-modal Prompt Encoder.}
We utilize the tokenizer and word embedding from pre-trained LLM~\cite{opt/zhang2022opt,llama2/touvron2023llama, vicuna/chiang2023vicuna} to process text and coordinate instructions. 
For image inputs, we employed the pre-trained ViT-L/14~\cite{clip/radford2021learning} as the image encoder, adopting a trainable projector based on the LLaVA~\cite{llava/liu2023visual} to collect image tokens.
Regarding 3D object inputs, we utilized a pre-trained 3D encoder~\cite{dc_vote2cap/chen2023end} to extract object features, obtaining object-level tokens via another projector. 
For point-level and box-level prompts, we directly employ linear layers to project the corresponding prompt features into the LLM embedding space. 
We leave the design of more optimal models for exploration in future work.

\myparagraph{Trainable Parameters.}
The model comprises roughly 52 million trainable parameters, accounting for less than 1\% of the frozen LLM backbone's (LLaMA-2-7B~\cite{llama2/touvron2023llama}) parameter count.

\section{GPT-4 Evaluation}
\label{app:gpt-4}

\begin{figure*}[htbp]
	\centering
	\includegraphics[width=0.9\linewidth]{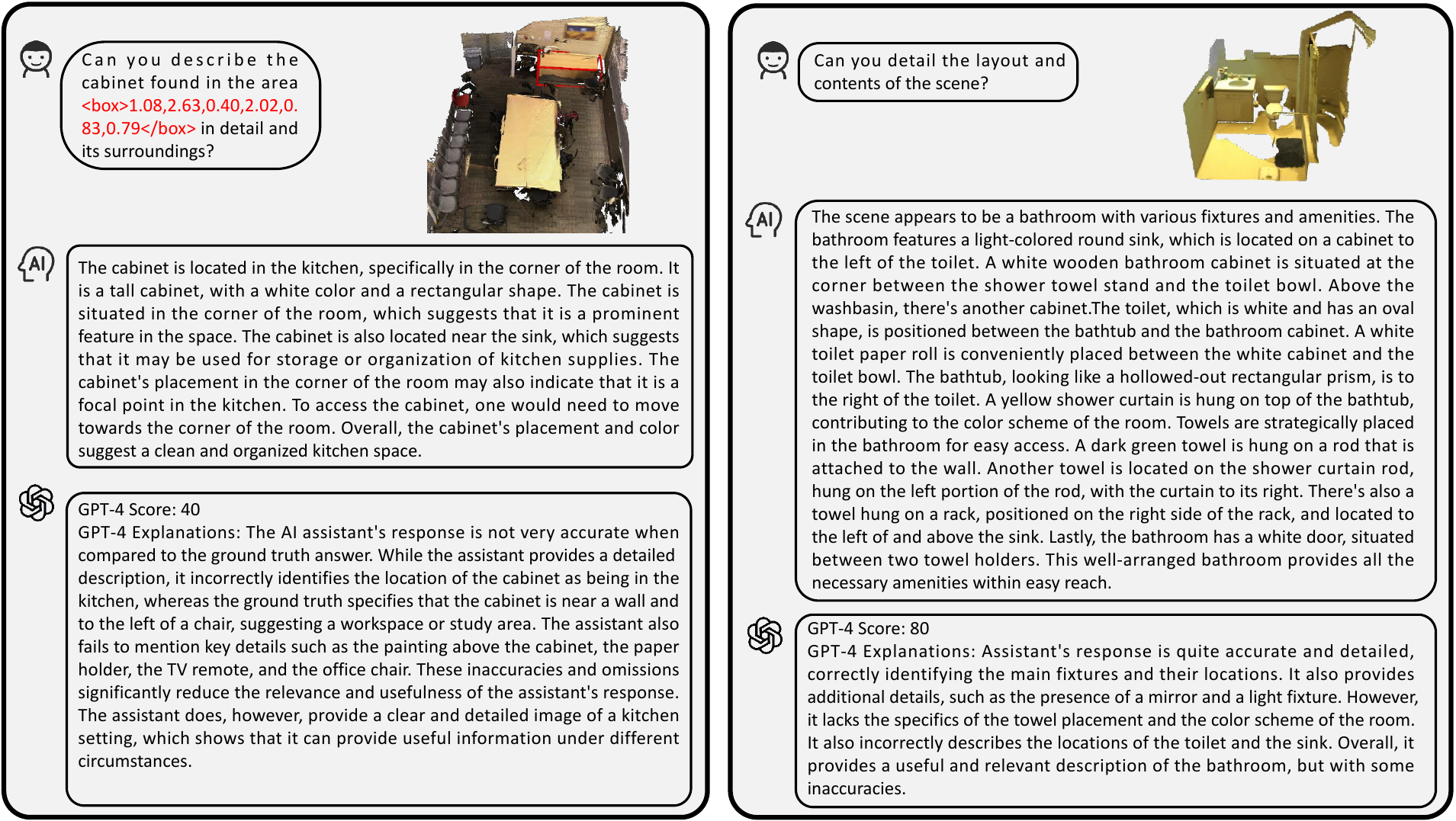}
	\caption{
        \textbf{Qualitative results on scene description.} We present visualization results for scene description, along with the scores provided by GPT-4 and the criteria used by GPT-4 for scoring.
        }
	\label{fig:qual_sd}
    \vspace{5pt}
\end{figure*}

\begin{figure*}[htbp]
	\centering
	\includegraphics[width=0.9\linewidth]{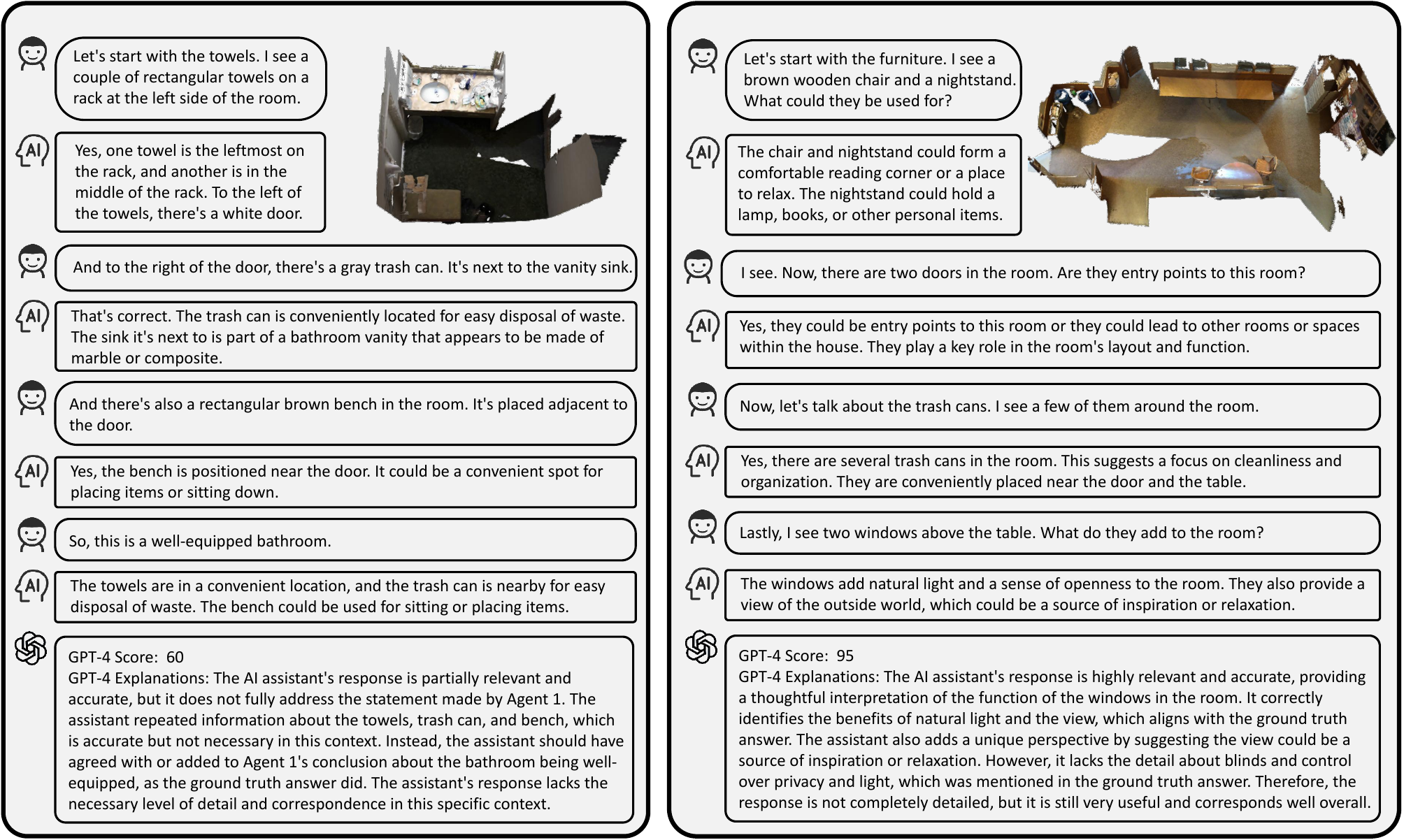}
	\caption{
        \textbf{Qualitative results on multi-round dialogue.} We present visualization results for multi-round dialogue, along with the scores provided by GPT-4 and the criteria used by GPT-4 for scoring.
        }
	\label{fig:qual_md}
    \vspace{5pt}
\end{figure*}

We employ the template in \cref{table:gpt4_prompt} to prompt GPT-4~\cite{gpt-4/openai2023gpt} and obtain corresponding evaluation results. 
Specifically, we prompt GPT-4~\cite{gpt-4/openai2023gpt} with four inputs: system message, question, reference answer, and models' responses, which comprise answers from various baseline models.
We prompt GPT-4~\cite{gpt-4/openai2023gpt} to assess responses for accuracy, relevance, descriptive details, etc., and assign scores within a range of 0 to 100. Higher scores indicate better quality of the responses.
Moreover, we request GPT-4~\cite{gpt-4/openai2023gpt} to provide explanations for the scoring results, assisting in our evaluation of the scoring results' validity. 
In \cref{fig:qual_sd} and \cref{fig:qual_md}, we present model's (utilizing transformer~\cite{transformer/vaswani2017attention} as the scene encoder and LLaMA~\cite{llama2/touvron2023llama} as the language decoder) responses, GPT-4 scores, and GPT-4 justifications on detailed description and multi-round dialogue.

\begin{table*}[htbp]
        \vspace{10pt}
	\centering
	\includegraphics[width=0.9\linewidth]{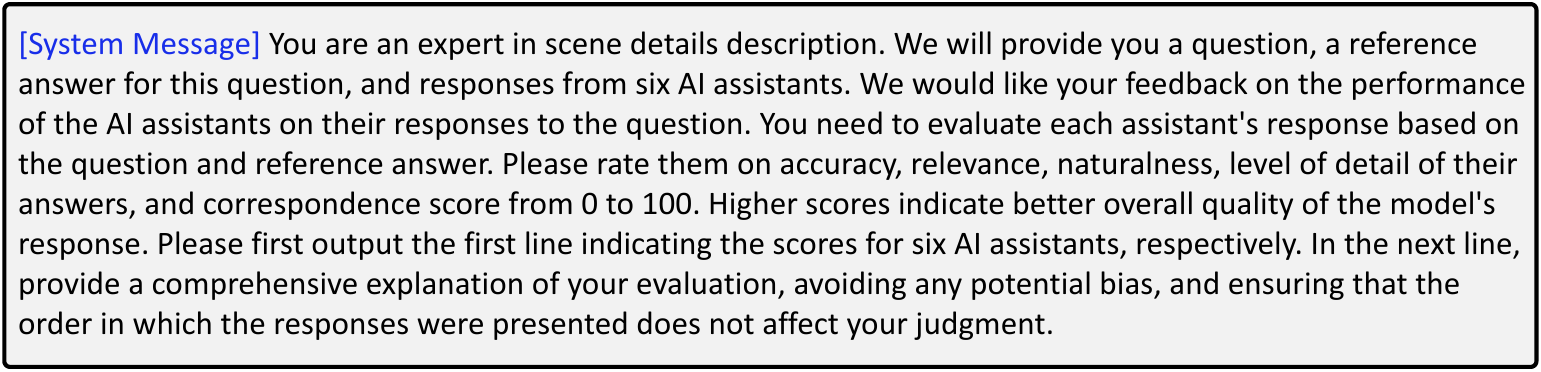}
	\caption{
        The template employed to prompt GPT-4 for assessing the quality of responses across different models.
        }
	\label{table:gpt4_prompt}
    \vspace{5pt}
\end{table*}

\newpage




\end{document}